\def\arxiv{1}

\if\arxiv1
\documentclass[final,5p,times,twocolumn]{ourarticle}
\else
\documentclass[final,5p,times,twocolumn]{elsarticle}
\fi

\usepackage{algorithm}
\usepackage{algorithmic}
\usepackage{amsfonts}
\usepackage{amsmath}
\usepackage{amssymb}
\usepackage{amsthm}
\usepackage{bm}
\usepackage{color}
\usepackage{enumitem}
\usepackage{epstopdf}
\usepackage{graphicx}
\usepackage{mathrsfs}
\usepackage{natbib}
\usepackage{subfigure} 
\usepackage{tikz}

\if\arxiv1
\usepackage{times}
\usepackage{txfonts}
\fi

\usetikzlibrary{arrows,calc,shapes,snakes,automata,backgrounds,fit,petri}

\allowdisplaybreaks[1]

\graphicspath{{figures/}}

\def\bchi{\bm{\chi}}

\def\bxi{\bm{\xi}}

\def\0{\mathbf{0}}
\def\1{\mathbf{1}}
\def\a{\mathbf{a}}
\def\b{\mathbf{b}}

\def\h{\mathbf{h}}

\def\s{\mathbf{s}}

\def\x{\mathbf{x}}
\def\y{\mathbf{y}}
\def\z{\mathbf{z}}

\def\I{\mathbf{I}}

\def\K{\mathbf{K}}

\def\S{\mathbf{S}}

\def\W{\mathbf{W}}

\def\drm{\mathrm{d}}
\def\Lcal{\mathcal{L}}

\def\Ncal{\mathcal{N}}
\def\Ocal{\mathcal{O}}

\def\Ebb{\mathbb{E}}

\def\Vbb{\mathbb{V}}

\def\wo{\backslash}
\def\defined{\stackrel{.}{=}}%{\stackrel{\text{\tiny def}}{=}}
\newcommand{\appropto}{\mathrel{\vcenter{
  \offinterlineskip\halign{\hfil$##$\cr
    \propto\cr\noalign{\kern2pt}\sim\cr\noalign{\kern-2pt}}}}}

\definecolor{Blue}{rgb}{0.0,0.0,1.0}

\newtheorem{theo1}{Theorem}
\newtheorem{prop1}{Proposition}

\journal{Digital Signal Processing}

\begin{document}

\begin{frontmatter}

\title{An Adaptive Resample-Move Algorithm for Estimating Normalizing Constants}

\author[dtu]{Marco Fraccaro}
\ead{marfra@dtu.dk}

\author{Ulrich Paquet}
\ead{ulrich@cantab.net}

\author[dtu]{Ole Winther\corref{corresponding}}
\ead{olwi@dtu.dk}

\cortext[corresponding]{Corresponding author}
\address[dtu]{Technical University of Denmark, Lyngby, Denmark}
% \address[up]{Cambridge, United Kingdom}

%!TEX root = main.tex

\begin{abstract} 
The estimation of normalizing constants is a fundamental step in probabilistic model comparison.
Sequential Monte Carlo methods may be used for this task and have the advantage of being inherently parallelizable.
However, the standard choice of using a fixed number of particles at each iteration is suboptimal because some steps will contribute disproportionately to the variance of the estimate.
We introduce an adaptive version of the Resample-Move algorithm, in which the
particle set is adaptively expanded
whenever a better approximation of an intermediate distribution is needed. 
The algorithm builds on 
the expression for the optimal number of particles and the corresponding minimum variance found under ideal conditions.
Benchmark results on challenging Gaussian Process Classification and Restricted Boltzmann Machine
applications show that Adaptive Resample-Move (ARM)
estimates the normalizing constant with a smaller variance, using less computational resources,
than either Resample-Move with a fixed number of particles or Annealed Importance Sampling.
A further advantage over Annealed Importance Sampling is that ARM is easier to tune.
\end{abstract} 


\end{frontmatter}

%!TEX root = main.tex

\section{Introduction}\label{sec:intro}

Any probabilistic model over random variables $\x$ can be framed as a nonnegative function
$f(\x)$ over the support of $\x$, that, when appropriately normalized by a partition function or normalizing constant $Z = \int f(\x) \, \drm \x$,
gives a probability density $p(\x) = \frac{1}{Z} f(\x)$.
In many cases $Z$ cannot be analytically evaluated, and needs to be numerically approximated.
The approximation can take a deterministic or stochastic form.
Deterministic methods turn the integration (or summation) required to obtain $Z$ into
optimization problems, sometimes through tightening a bound on $Z$ \cite{jordan1999introduction},
but introduce an unknown approximation error.
Stochastic methods that approximate $Z$ through a Monte Carlo estimate are exact in the infinite-sample limit
\cite{robert2004montecarlo}, but unlike their deterministic counterparts, are more expensive to tune and run \cite{Gelman2013}.
When $p(\x)$ is decomposed into a sequence of distributions, particle filters or sequential Monte Carlo (SMC) methods
track the sequential changes in the probability mass that $p$ assigns to the domain on $\x$
\cite{halton1962sequential}.
SMC's idea of propagating a set of particles that iteratively describe the changing probability mass, 
forms the foundation of many efficient tools for statistical inference \cite{introSMC}.

SMC's success relies on inner details, like how to use intermediate transition kernels to
navigate the changing probability mass as efficiently as possible, or simply how many particles to use.
How to maximally use computational resources in SMC algorithms is an open research problem,
with recent efforts focusing on parallelized and distributed implementations \citep{PaiWooDou2014a,Whiteley2014}.
In this work we approach
this issue from a different perspective, showing that SMC methods can also
be made more efficient by using less particles, if they are used optimally.
While a few particles are needed to provide an accurate empirical estimate of an intermediate distribution in SMC's sequential decomposition,
in other ``high variance'' iterations more reliable results can be obtained if the particle set is extended to better approximate expectations of interest.
Our proposed algorithm, \textit{Adaptive Resample-Move}, represents a theoretically grounded way to exploit this idea for optimizing the estimation of normalizing constants.

Under a fixed computational budget,
the optimal way to minimize the variance of the estimate of the normalizing constant is to use a variable number of particles at each iteration.
This is proved in Section \ref{sec:rm} for the ideal condition of independent samples.
This result is then used to define \textit{Adaptive Resample-Move} (ARM),
an extension of an SMC method known as Resample-Move \citep{ResampleMove}.
ARM finds accurate estimates using an adaptive number of particles at each iteration (see Section \ref{sec:rmg}).
Experimentally we show that, from a computational view, it is better to adaptively grow the
number of particles per iteration as needed.

The proposed algorithm is compared to state of the art methods on two sets of challenging machine learning problems:
In Section \ref{sec:gpc}, ARM is compared to several versions of Annealed Importance Sampling \cite{NealAIS} for Gaussian Process (GP) Classification.
To obtain a competitive baseline,
we derived a Riemannian Manifold Hamiltonian Monte Carlo \citep{RMHMC} sampler for GP models,
which would be of independent interest \citep{paquet16riemannian}.
Due to the importance of Restricted Boltzmann Machines to the deep learning community, we evaluated 
ARM in Section \ref{sec:rbm} to estimate their normalizing constants.
The results indicate that ARM provides very competitive accuracy at a lower computational cost and perhaps most importantly with more ease for adapting the interpolation to the problem at hand.
There exists a large body of related work, which we describe in the context of this paper in
Section \ref{relWork}.

%!TEX root = main.tex

\section{The Resample-Move algorithm} \label{sec:rm}

Sequential Monte Carlo (SMC) algorithms \citep{introSMC}
obtain samples from a target distribution $p$ by iteratively sampling from a sequence of distributions $p_1, p_2, \dots, p_N = p$.
Although SMC is more generally applicable, we restrict ourselves to a sequence of distributions
\[
p_n(\x_{[n]}) = \frac{1}{Z_n} f_n(\x_{[n]})
\]
that are defined on spaces
of increasing dimensionality, where $\x_{[n]} = (x_1, x_2, \ldots, x_n)$ indicates the vector of the first $n$ components of $\x$, and 
$$
Z_n = \int f_n(\x_{[n]}) \, \drm \x_{[n]} \ .
$$
Note that each consecutive $f_n$ has \emph{no} dependence on variables $x_i$ for $i > n$.

At iteration $n$, where the iterations run $n = 1, \ldots, N$, a set of $R$ particles
$\x^{[R]}_{[n]} = (\x^{1}_{[n]}, \x^{2}_{[n]}, \ldots, \x^{R}_{[n]})$ with weights
$w^{[R]}_{n} = (w_n^1, w_n^2, \ldots, w_n^R)$ are kept,
such that they provide an empirical estimate of $p_n$, in the sense that
\begin{equation}\label{eq:target}
\textstyle{
\sum_{r=1}^R \widetilde{w}_n^r \, \varphi( \x_{[n]}^r) \rightarrow \Ebb_n \big[ \varphi(\x_{[n]}) \big] 
}
\end{equation}
almost surely as $R\rightarrow \infty$, for any measurable $\varphi$ such that the expectation $\Ebb_n[\varphi(\x_{[n]})]$ exists.
Notation
$\widetilde{w}_n^r = w_n^r / \sum_{r'=1}^R w_n^{r'}$
indicates the normalized weight of the $r$'th particle, and
$\Ebb_n$ is a shorthand for the expectation $\Ebb_{p_n(\x_{[n]})}$.
If \eqref{eq:target} holds, we say that $(\x^{[R]}_{[n]}, w^{[R]}_{n})$ \emph{targets} $p_n(\x_{[n]})$.
The target of the particle system evolves over time: samples from $p_{n+1}(\x_{[n+1]})$ are obtained with importance sampling and resampling techniques using $p_{n}(\x_{[n]})$ as a proposal distribution.

\begin{figure}[t]
\begin{center}
\includegraphics[width=0.42\textwidth]{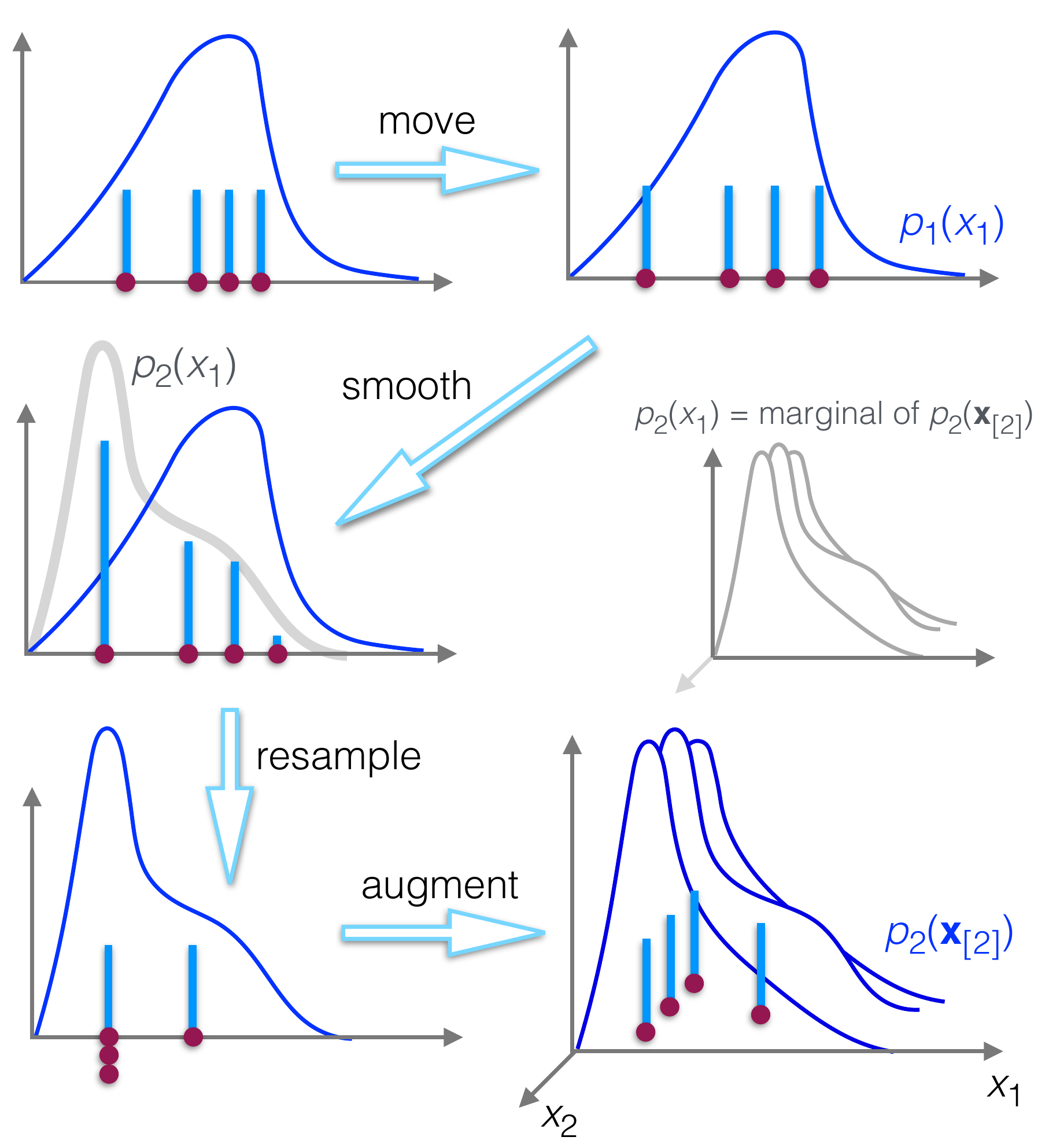}
\caption{A sketch of the first iteration of Resample-Move in Algorithm \ref{alg:rm},
starting with four weighted particles $x_{1}^{[4]}$ that target $p_1(x_1)$.
The heights of the bars reflect the particle weights $w_1^{[4]}$ (top left).
After application of a MCMC transition kernel ${\cal K} ( x_{1} ; x^{[4]}_{1})$ in the \textbf{move}-step, the moved particles $x^{[4]}_{1}$ still target $p_1(x_1)$.
In the \textbf{smooth}-step, the weights of the particles are adjusted to
$w_2^{[4]}$, so that they target the marginal distribution $p_2(x_1) \defined \int p_{2}(\x_{[2]}) \, \drm x_{2}$.
The implication of this step is that one can extend each particle by a new dimension in the
\textbf{augment}-step, by sampling $x_2^r$ from $p_2(x_2 | x_1^r)$ 
for $r=1, \ldots, 4$.
Their weights are kept unchanged.
The newly augmented (weighted) particles target $p_2(\x_{[2]})$.
Before the augment-step, one might want to reset the weights $w_2^{[4]}$ on the set $x_1^{[4]}$
to be uniform.
This is done by stochastically replicating high-weighted particles under $p_2(x_1)$ in the \textbf{resample}-step. 
%The move-step in the next iteration will then help to increase the diversity in the replicated particles.
}
\label{fig:illustration}
\end{center}
\end{figure}

The normalizing constant $Z = Z_N$ of the target distribution unrolls over the sequence
with
\begin{align} 
\log Z &
= \log Z_1 + \sum_{n=1}^{N-1} \log \frac{Z_{n+1}}{Z_n} \nonumber \\
& = \log Z_1 + \sum_{n=1}^{N-1} \log \frac{1}{Z_n}  \int \frac{ f_{n+1} \left( \x_{[n+1]} \right) }{
f_n \left( \x_{[n]} \right) } f_n(\x_{[n]})  \, \drm \x_{[n+1]} \nonumber \\
& = \log Z_1 + \sum_{n=1}^{N-1} \log \Ebb_n \left[ \int \frac{ f_{n+1} \left( \x_{[n+1]} \right) }{
f_n \left( \x_{[n]} \right) } \, \drm x_{n+1} \right] .
\label{eq:decomposition}
\end{align}
A recursive unbiased estimate of $\log Z$ can then be obtained using the set of particles
that target $p_n(\x_{[n]})$ to approximate the expectation in \eqref{eq:decomposition} with a weighted average. Starting at $x_1$, the number of variables averaged over is therefore sequentially increased by one with the Resample-Move algorithm outlined in Algorithm \ref{alg:rm}, which uses any random ordering of variables to decompose $\log Z$. The algorithm's key steps are illustrated in Figure \ref{fig:illustration} and discussed in detail in Section \ref{subsec:rm}.

\begin{algorithm}[tb]
\begin{algorithmic}[1]
\STATE $x^{[R]}_{1} \sim p_1(x_{1})$;
$\; \widetilde{w}^{[R]}_{1} := \frac{1}{R}$;
$\; \log Z := \log Z_1$
\FOR {$n=1$ to $N-1$}
	\STATE $\x^{[R]}_{[n]} \sim {\cal K} \big( \x_{[n]} \, ; \, \x^{[R]}_{[n]} \big)$
\hfill \emph{// move} 
%%\label{alg:rm:move}
	\STATE $w^{[R]}_{n+1} := \mathcal{W} \big( \x^{[R]}_{[n]}\big) \, \widetilde{w}^{[R]}_n$
\hfill \emph{// smooth} 
	\STATE $\log Z := \log Z + \log \sum_r w^{r}_{n+1}$  
	\STATE $\widetilde{w}^{[R]}_{n+1} := w^{[R]}_{n+1}/\sum_i w^{i}_{n+1}$
	\IF {$R_{{\rm eff}} \big( \widetilde{w}_{n+1}^{[R]} \big) < R_{{\rm eff}}^{\min}$}
	\STATE $\x^{[R]}_{[n]} := {\rm resample} \big (\widetilde{w}^{[R]}_{n+1},\x^{[R]}_{[n]} \big)$
\hfill \emph{// resample}
	\STATE $\widetilde{w}^{[R]}_{n+1} := \frac{1}{R}$
	\ENDIF
	\STATE $x^{[R]}_{n+1} \sim x_{n+1} | \x^{[R]}_{[n]}$
\hfill \emph{// augment}
\ENDFOR 
\STATE \textbf{return} $\log Z$ and $\x^{[R]}_{[N]}$
\end{algorithmic}
\caption{Resample-Move}
\label{alg:rm}
\end{algorithm}

\subsection{Resample-Move}\label{subsec:rm}

Resample-Move (RM) \citep{ResampleMove} extends standard SMC methods using an MCMC kernel to increase diversity in the particles (see also Section~\ref{sec:shortcomingsSMC}).
It can be seen as a special case of the very general SMC framework introduced in \citep{smcsamplers}.
As invariant pre-condition to Line 3's move-step, $(\x_{[n]}^{[R]}, \widetilde{w}_n^{[R]})$ targets $p_n$, and this property is retained
after application of a Markov chain Monte Carlo (MCMC) transition kernel ${\cal K}$ that has $p_n(\x_{[n]})$ as invariant distribution.
The smooth-step in Line 4 computes the smoothed ratio
\begin{equation} \label{eq:smooth}
\mathcal{W} \big( \x_{[n]} \big) = \frac{p_{n+1}(\x_{[n]})}{p_{n}(\x_{[n]})} = \int \frac{ f_{n+1} ( \x_{[n+1]} )  }{ f_n ( \x_{[n]} ) } \, \drm x_{n+1},
\end{equation}
by marginalizing out $x_{n+1}$.
Notation
$p_{n+1}(\x_{[n]}) \defined \int p_{n+1}(\x_{[n+1]}) \, \drm x_{n+1}$ denotes the \emph{marginal} distribution of $p_{n+1}(\x_{[n+1]})$.
With the updated importance weights being $w^{r}_{n+1} = \mathcal{W} (\x^{r}_{[n]}) \widetilde{w}^{r}_n$,
 the set $(\x^{[R]}_{[n]}, \widetilde{w}_{n+1}^{[R]})$ will target $p_{n+1}(\x_{[n]})$.
Note that $\mathcal{W}( \x_{[n]} )$ is exactly the argument of the expectation in
\eqref{eq:decomposition}; it is hence possible to update the incremental estimate of $\log Z$
(Line 5) using the Monte Carlo approximation in \eqref{eq:target}.
To ensure that Line 3's precondition will hold in the next iteration,
the augmentation-step in Line 11
adds a component $x_{n+1}^{r}$ to each particle by sampling from the conditional
distribution $x_{n+1}^{r} \sim p_{n+1}(x_{n+1} | \x_{[n]}^{r})$.
The particles $(\x^{[R]}_{[n+1]},\widetilde{w}_{n+1}^{[R]})$ will then target $p_{n+1}(\x_{[n+1]})$.

The resample-step in Lines 8 and 9 is wedged between the smooth- and augmentation-steps,
and returns a new set of uniformly weighted particles that are resampled from the old ones
using weights $\widetilde{w}_{n+1}$. Multinomial or residual resampling \citep{LiuChenSMC} 
are commonly used.\footnote{Asymptotically, residual resampling will always outperform multinomial resampling \citep{Chopin2004}.}
After Line 9, the set still targets $p_{n+1}(\x_{[n]})$, and is formed in such a way that the most significant particles are repeated to serve as multiple starting points for the next move-step,
while particles with low weights are discarded.

Resampling should be done only if necessary to prevent the \emph{degeneracy} problem
(that is, when $\widetilde{w}_{n+1}^{[R]}$ is such that only few particles have a significant weight,
and all the others have a very small weight), as resampling introduces correlation among particles and additional variance in the estimates. It is common to use the \emph{Effective Sample Size} (ESS) \cite{LiuChenSMC}
\begin{equation} \label{ess}
R_{{\rm eff}} \big( w_{n+1}^{[R]} \big)
= \frac{ ( \sum_{r=1}^{R} {w}^r_{n+1} )^2 } {\sum_{r=1}^{R} ( {w}^r_{n+1} )^2 }
= \frac{1}{\sum_{r=1}^{R} ( \widetilde{w}^r_{n+1} )^2 }
\end{equation}
as a yardstick to measure the number of particles with a significantly high weight.
The resample-step is then done only if the ESS is below a certain threshold, for example $R_{{\rm eff}}^{\min} = 0.7R$.

%% WHEN DOES RM FAIL?
%degeneracy , hence resample hence sample impoverishment
%
%in RM, R e M steps are inserted to prevent sample impoverishment (http://www.cns.nyu.edu/~eorhan/notes/particle-filtering.pdf)
%however it may still fail\\
%
%1. when two seq distr are too different (this issue is common to all smc methods) \\
%
%a- only few of the particles after the move step are important in the new step (similar ideas in auxiliary variable SMC, put it in literature review! http://www.stats.gla.ac.uk/~levers/teaching/mcm/slides15-2x2.pdf)\\
%
%b- lost track, i.e. none of the particle is important, they approximate a useless region - but in this case ESS is misleading\\
%
%2. degeneracy problem not solved (diversity not increased): slow mixing kernel not run long enough - many "almost copies" of the same particle/ we don't explore enough
%
%% WHAT'S THE ISSUE IF IT FAILS?
%
%1a: estmate of log(Zn+1/Zn) based only on few important points (low ess)\\
%
%1b: estmate of log(Zn+1/Zn) based only on many useless points (misleading high ess)\\
%
%2. we may have all the particles in the high density region but they do not approximate it properly, as they are too similar (misleading high ess)

\subsection{Shortcomings of RM and other SMC algorithms}\label{sec:shortcomingsSMC}

The resample-step is a powerful way to deal with the degeneracy problem, as after the resample-step,
all particles have an equal weight. However, it introduces a new issue, namely \textit{sample impoverishment}. 
Particles with a high weight are likely to be resampled many times, and this means that the
actual number of particles contributing to the weighted average in \eqref{eq:target} may be much smaller than $R$.
RM reduces sample impoverishment with the move-step,
which increases diversity in the particles.
These additional steps are useful only if any two consecutive sequential distributions are similar enough: in our case ${p_{n}(\x_{[n]})}$ has to be reasonably close to $p_{n+1}(\x_{[n]})$; see \eqref{eq:smooth} and Figure \ref{fig:illustration}.
If they differ too much, only a few particles might suddenly be significant in the next iteration (in the worst case no particles would fall in the high probability density region),
making it very difficult for the system to provide again a good approximation of any expectation of interest.
In sequential parameter estimation, for example, this could happen when a particularly difficult
data observation is to be introduced \citep{chopin}, and the distribution
changes to one that is very different.
Section~\ref{sec:theo_opt} proves a theorem that under % ideal conditions and
%% Marco - I've removed the ideal conditions precondition here, as we mentioned it in the introduction, and at the start of the theorem.
a fixed computational budget $R_{{\rm tot}}$,
the difference between distributions translates to RM requiring a higher number of particles
$R_n$ in iteration $n$.
% In RM problems may also occur in the presence of a slow mixing kernel that is
% not run long enough in the move-step. In this case the sample impoverishment problem would not be solved, and we would have very similar copies of the resampled particles.

\subsection{Optimal number of particles at each iteration}\label{sec:theo_opt}

We argued that if $p_{n}(\x_{[n]})$ and $p_{n+1}(\x_{[n]})$ are not similar enough, then a higher number of particles is needed, and formalize the statement here.
Let the number of particles at each iteration be variable,
so that $p_n (\x_{[n]})$ is approximated with $R_n$ particles.
Given a computational budget of
 $R_{{\rm tot}} = \sum_{n=1}^{N-1} R_n$, we may wonder what the optimal values for $R_n$ for $n=1,\dots N-1$ are, such that the variance of the estimate of the normalizing constant
\begin{equation} \label{eq:decomposition_estimator}
% \log \widehat{Z} = \log Z_1 + \sum_{n=1}^{N-1} \log \left( \sum_{r=1}^{R_n}  \mathcal{W} \big( \x^{r}_{[n]} \big) \widetilde{w}^{r}_n \right) 
\textstyle{ \log \widehat{Z} = \log Z_1 + \sum_{n=1}^{N-1} \log \big( \sum_{r=1}^{R_n}  \mathcal{W} ( \x^{r}_{[n]} ) \widetilde{w}^{r}_n \big) }
\end{equation}
is minimized (see \eqref{eq:decomposition}). The following theorem gives the answer to this question under ideal conditions.
Let $\Vbb_n[\cdot]$ denote the variance of its argument under $p_n(\x_{[n]})$.

\begin{theo1}\label{theo:1}
Assume independent equally-weighted samples from the distributions $p_n(\x_{[n]})$ for $n=1,\dots N-1$, and define the variance of the normalized weight updates
\[
v_{n} \doteq
\Vbb_n \left[ \frac{\mathcal{W} ( \x_{[n]} )}{\Ebb_n [ \mathcal{W}(\x_{[n]}) ]} \right] \ .
\]
The optimal values for $R_1, \dots, R_{N-1}$ that minimize the variance of the estimate
$\log \widehat{Z}$ from \eqref{eq:decomposition_estimator} are
\begin{equation}\label{eq:Ropt}
R_n^{\rm opt} =  \frac{\sqrt{v_{n}}}{\sum_{n^\prime=1}^{N-1} \sqrt{v_{n^\prime}}} \, R_{\rm tot} \ .
\end{equation}
% where $R_{tot}=\sum_{n=1}^{N-1} R_n$.
The corresponding minimum variance of $\log \widehat{Z}$ is
\begin{equation}\label{eq:Vmin}
\textstyle{ V_{\min} = \big( \sum_{n=1}^{N-1} \sqrt{v_{n}} \big)^2 / R_{\rm tot} } \  .
\end{equation}
\end{theo1}

\begin{proof} 
Due to the independence of the samples, the variance that we want to minimize can be decomposed as
$$
\Ebb_1 \ldots \Ebb_{N-1} \left[ ( \log \widehat{Z} - \log Z)^2 \right]  = \sum_{n=1}^{N-1} \Vbb_n \left[\log m_{n}\right] \ , 
$$
where $m_{n} = \frac{1}{R_n} \sum_{r=1}^{R_n} \mathcal{W} ( \x_{[n]}^r )$ represents a sample average.
For large $R_n$, the central limit theorem implies that $m_{n}$ converges in distribution to a Gaussian
$\Ncal \big( \Ebb_n [\mathcal{W} ( \x_{[n]} )], \,
\frac{1}{R_n} \Vbb_n [ \mathcal{W} ( \x_{[n]} ) ] \big)$, hence the delta method can be used to approximate $\log m_{n}$:
$$
\log m_{n} \approx \Ncal\left(\log \Ebb_n [\mathcal{W} ( \x_{[n]} )], \, 
\frac{v_{n}}{R_n} \right) \ .
$$
Having found an expression for $\Vbb_n [ \log m_{n} ]$, the constrained optimization problem can therefore be rewritten as
$$
\begin{aligned}
& \underset{R_1,..R_{N-1}}{	\min}
& & \sum_{n=1}^{N-1} \frac{v_{n}}{R_n}
& \text{subject to}
& & \sum_{n=1}^{N-1} R_n=R_{\rm tot} \\
\end{aligned} \ ,
$$
and can be solved using Lagrange multipliers to find, after some calculations, the results stated in the theorem.
\end{proof} 
Theorem \ref{theo:1} implies that, given a fixed computational budget,
a variable number of particles has to be used per iteration
if the variance of $\log \widehat{Z}$ is to be minimized. From \eqref{eq:Ropt}
we deduce that, as expected, when the variance $v_{n}$ of the normalized weight updates is big, i.e.~if ${p_{n}(\x_{[n]})}$ and ${p_{n+1}(\x_{[n]})}$ are not similar enough, then a higher number of particles is required.

\paragraph{Adaptively adding particles per iteration}
Assuming that the computational budget can be exceeded in iteration $n$,
how can particles be added adaptively so that $V_{\min}$ is decreased?
We can glean some insight by considering the contribution of iteration $n$ to $V_{\min}$.
Firstly, we obtain an approximation to $v_{n}$ with its empirical estimate
\[
\widehat{v}_n = \frac{\frac{1}{R_n} \sum_r (\mathcal{W} ( \x^r_{[n]} )- m_n)^2}{m_n^2 }  = \frac{R_n}{R_{{\rm eff},n}} - 1 \ ,
\]
where $R_{{\rm eff},n} \defined R_{{\rm eff}}([w_{n+1}^{[R_n]}])$.
Assuming that $R_i = R_i^{{\rm opt}}$ for all iterations,
the contribution of iteration $n$ can be isolated in $V_{\min}$
by substituting all $\widehat{v}_i$ into \eqref{eq:Vmin}:
\begin{equation}\label{eq:VminApprox}
V_{\min} \approx \frac{ \Big( \sqrt{ R_n^{\rm opt} / R_{{\rm eff},n}^{\rm opt} - 1} +
\sum_{i \neq n} \sqrt{ R_i^{\rm opt} / R_{{\rm eff},i}^{\rm opt} - 1} \Big)^2
}{
R_n^{\rm opt} + \sum_{i\neq n}R_i^{\rm opt}} \  .
\end{equation}
If we are now allowed to exceed the computational budget, 
we can visualize one possible way to further decrease the variance $V_{\min}$.
If at iteration $n$ we increased the number of particles
to $R_n^* > R_n^{\rm opt}$ particles that still target
$p_{n}(\x_{[n]})$, then $V_{\min}$ could be decreased provided that
$R_{{\rm eff},n}^{\rm opt} / R_n^{\rm opt}
< R^*_{{\rm eff},n} / R^*_n \leq 1$.
The variance decreases when particles are added so that the ESS \emph{per particle}
increases.\footnote{Looking at the ESS alone is not sufficient. As a simple argument,
the ESS in (\ref{ess}) can be doubled by simply duplicating each particle, but this
doesn't alter the ESS per particle.}
This intuition represents the starting point for the sampler that is developed next.

%!TEX root = main.tex

\section{Adaptive Resample-Move}\label{sec:rmg}

Theorem \ref{theo:1} dictates how to optimally divide a fixed particle budget
if the variances $v_n$ of $w_{n+1} / \Ebb_n[w_{n+1}]$
are known under i.i.d.~conditions for $n = 1, \ldots, N-1$.
With Algorithm \ref{alg:rm} being sequential and having no knowledge of future iterations $n' > n$,
we can greedily try to keep $R_{\rm tot}$ small, by using the ESS as a gauge for adaptively setting $R_n$.
At a high level, iteration $n$ starts with $R_n = R$ particles, and while a condition based on the ESS is not met, the iteration's number of particles is increased to $R_n := R_n + R$ through various means,
as explained in Section \ref{sec:gener} (at most $i_{\max}$ times).
This ensures $\frac{1}{N-1}R_{\rm tot} \le i_{\max} R$, although experimentally the average number
of particles is much smaller, with $\frac{1}{N-1}R_{\rm tot} \approx 1.5 R$ for Section \ref{sec:gpc}'s results.
We introduce this adaptive ``generate'' loop to the RM in Algorithm \ref{alg:rmg},
and call it \textit{Adaptive Resample-Move} (ARM).
Note that only few lines in Algorithm \ref{alg:rm} need to be changed.
Whenever a better approximation of the probability distribution is needed, ARM generates new particles that target it.
At the other extreme end, if all components of $\x$ are independent, it captures the fact
that no application of a move-step transition kernel would ever be required.

% In the analysis in Section \ref{sec:shortcomingsSMC}, we have seen that the main problems of SMC methods are due to bad approximations given by the particles when two consecutive sequential distributions are not similar enough.
% In sequential parameter estimation for example \citep{chopin}, this could happen when a particularly difficult observation is introduced.
% As to minimize the variance of the estimate of the normalizing constant  a variable number of particles per iteration is optimal (see Section \ref{sec:theo_opt}), we propose to add a further step in the RM algorithm.  Whenever a better approximation of the probability distribution is needed, new particles that approximate the target distribution are generated. We call this algorithm \textit{Adaptive Resample-Move} (ARM).

%We will now introduce intuitively a principled approach to extend the particles set, leaving the more theoretical discussion on the asymptotic properties of the estimators to Section \ref{sec:asympt}. Ways to generate new particles will be described in Section \ref{sec:gener}.

We next present two sufficient conditions that allow us to enlarge the particle set at any iteration $n$.

\begin{prop1}\label{prop1}
Let $(\bchi^{[R_n]}_{[n]},\widetilde{\rho}_n^{[R_n]})$ be a set of $R_n$ particles that
target $p_{n}(\x_{[n]})$. If we have $R$ new particles $(\x^{[R]}_{[n]},\widetilde{w}_{n}^{[R]})$ such that
% \vspace*{-0.25cm}
\begin{enumerate}
\item $(\x^{[R]}_{[n]},\widetilde{w}_{n}^{[R]})$ targets $p_{n}(\x_{[n]})$, and
\item the $R_n + R$ particles' weights are rescaled as $\alpha \widetilde{\rho}_n^{[R_n]}$ and $(1 - \alpha) \widetilde{w}_{n}^{[R]} $, with $0 \le \alpha \le 1$,
\end{enumerate}
%\setlist[description]{font=\normalfont}
%\begin{description}
%  \setlength{\itemsep}{0pt}
%   \setlength{\parskip}{0pt}
%\item[Property 1:] $(\x^{[R]}_{[n]},\widetilde{w}_{n}^{[R]})$ targets $p_{n}(\x_{[n]})$;
%\item[Property 2:] the $R_n + R$ particles' weights are rescaled as $\alpha \widetilde{\rho}_n^{[R_n]}$ and $(1 - \alpha) \widetilde{w}_{n}^{[R]} $, with $0 \le \alpha \le 1$,
%\end{description}
% \vspace*{-0.25cm}
%%
then  $\big( (\bchi^{[R_n]}_{[n]}, \x^{[R]}_{[n]}), \, (\alpha \widetilde{\rho}_{n}^{[R_n]}, (1 - \alpha) \widetilde{w}_n^{[R]}) \big)$ also targets $p_{n}(\x_{[n]})$.
This also holds if the new set $\x^{[R]}_{[n]}$ is moved using a transition kernel that leaves $p_{n}(\x_{[n]})$ invariant.
\end{prop1}

\begin{proof} 
As $(\bchi^{[R_n]}_{[n]}, \widetilde{\rho}_n^{[R_n]})$ and
$(\x^{[R]}_{[n]}, \widetilde{w}_n^{[R]})$ both target $p_{n}(\x_{[n]})$, we have
\begin{align*}
\textstyle{ \alpha \sum_{r=1}^{R_n} \varphi(\bchi_n^r) \widetilde{\rho}_n^r} & \longrightarrow \textstyle{\alpha \, \Ebb_n[\varphi(\x_{[n]})] } \\
\textstyle{ (1 - \alpha) \sum_{r=1}^R \varphi(\x_n^r) \widetilde{w}_n^r } & \longrightarrow
\textstyle{ (1 - \alpha) \Ebb_n[\varphi(\x_{[n]})]} \ .
\end{align*}
The first statement is verified by combining the two sums.
It is also possible to move the new particles $\x^{[R]}_{[n]}$ using a transition kernel that leaves $p_{n}(\x_{[n]})$ invariant, as they would still target $p_{n}(\x_{[n]})$; see \citep{ResampleMove} for details.
\end{proof}

\newlength{\myspace}
\newlength{\mytab}
\setlength{\myspace}{11pt}
\setlength{\mytab}{9pt}

%%%%%%%%%%%%%%%%%%%%%%%%%%%%%%%%%%%%%%%%%%%%%%
\begin{algorithm}[t!]
\caption{Adaptive Resample-Move}
\label{alg:rmg}
\begin{algorithmic}
\STATE \hspace{\myspace}$\vdots$
\STATE {\small{4: }}$w^{[R]}_{n+1} := \mathcal{W} \big( \x^{[R]}_{[n]} \big) \, \widetilde{w}^{[R]}_{n}$ \hfill \emph{// smooth}
\STATE \hspace{\myspace}\textbf{if} {$\gamma^{(0)}<\gamma_{\rm thr}$} \textbf{then}
\STATE \hspace{\myspace}\hspace{\mytab}$\bchi^{[R]}_{[n]} := \varnothing$; 
$\ \widetilde{\rho}^{[R]}_{n} := \varnothing$; $\ i := 0$
\STATE \hspace{\myspace}\textbf{end if}
\STATE \hspace{\myspace}\textbf{while} {$\gamma^{(i)}<\gamma_{\rm thr}$} and $i < i_{\max}$ \textbf{do} \hfill \emph{// generate}
\STATE \hspace{\myspace}\hspace{\mytab}$\x^{[R]}_{[n]} \sim {\cal K}\big( \x_{[n]} \, ; \, \x^{[R]}_{[n]} \big)$ \hfill \emph{// move} 
\STATE \hspace{\myspace}\hspace{\mytab}$\bchi^{[R_n]}_{[n]} := ( \bchi^{[R_n]}_{[n]}, \x^{[R]}_{[n]} )$ \hfill \emph{// include samples} 
\STATE \hspace{\myspace}\hspace{\mytab}$\widetilde{\rho}^{[R_n]}_{n} := ( \frac{R_n}{R_n + R} \widetilde{\rho}^{[R_n]}_{n}, \;\frac{R}{R_n + R} \widetilde{w}^{[R]}_{n})$
\STATE \hspace{\myspace}\hspace{\mytab}$w^{[R_n]}_{n+1} := \mathcal{W} \big( \bchi^{[R_n]}_{[n]} \big) \, \widetilde{\rho}^{[R_n]}_{n}$ \hfill \emph{// smooth}
\STATE \hspace{\myspace}\hspace{\mytab}$i := i + 1$
\STATE \hspace{\myspace}\textbf{end while}
\STATE \hspace{\myspace}$\vdots$
\STATE {\small{7: }}\textbf{if} {$\gamma^{(i)}<\gamma_{\rm thr}$} or $R_n > R$ \textbf{then}
\STATE {\small{8: }}\hspace{\mytab}$\x^{[R]}_{[n]} := {\rm resample} \big (\widetilde{w}^{[R_n]}_{n+1}, \bchi^{[R_n]}_{[n]} \big)$ \hfill \emph{// resample}
\STATE \hspace{\myspace}$\vdots$
\end{algorithmic}
\end{algorithm}

\subsection{Adding a \emph{generate}-loop}\label{sec:alg_rmg}

We exploit Proposition \ref{prop1} to expand the move-step into a generate-loop in Algorithm \ref{alg:rmg},
by starting with a base level of $R_n^{(0)}$ particles
that is large enough to get a reliable estimate of the effective sample size $R^{(0)}_{{\rm eff},n}$.
Defining
\[
\gamma^{(i)} = R^{(i)}_{{\rm eff},n} / R_n^{(i)} \ ,
\]
for generate-iteration $i$, we see from \eqref{eq:VminApprox} that $R^{(0)}_{{\rm eff},n}$ would contribute
approximately $\sqrt{1 / \gamma^{(0)} - 1}$
to the minimum variance (via the first square root term).
If $\gamma^{(0)}$ is below a threshold value $\gamma_{\rm thr}$,
and more particles are generated such that the ratio $\gamma^{(1)}$ is increased,
then, as suggested in Section \ref{sec:theo_opt},
the variance could be further decreased. 
This procedure is iterated until $\gamma^{(i)}>\gamma_{\rm thr}$, or until 
a maximum number of iterations $i_{\max}$ are reached.
The threshold should be as close as possible to one, as the contribution to the variance
in \eqref{eq:VminApprox} is $\sqrt{ 1 / \gamma_{\rm thr} - 1}$.
One should however bear in mind that a higher threshold gives a computationally more expensive
algorithm. In our experiments, a value of $\gamma_{\rm thr}=0.7$ gave a good trade-off. 

As a cautionary tale, the ESS may be misleading, as it could be high even if an important mode in the distribution is missed. However, it gives a practically useful measurement of the quality
of the approximation, and is hence commonly used in SMC methods \citep{smcsamplers}.
Additionally, Theorem \ref{theo:1} rested on an assumption of independent, equally-weighted samples to make the study of some asymptotic properties of the introduced sampler feasible.
In practice, the resample-step introduces correlations, and the particles may not
yet be at equilibrium after the application of ${\cal K}$.
However, as our results in the following sections suggest,
ARM allows us to reduce the variance of the estimate of normalizing constants,
even if these assumptions are not fully satisfied.

\subsection{Generating the particles}\label{sec:gener}

As long as Proposition \ref{prop1}'s properties are satisfied, any method could be used to generate new particles.
As shown in Algorithm \ref{alg:rmg}, {ARM} creates $R$ new particles by repeating and moving the old set $(\x^{[R]}_{[n]}, \widetilde{w}^{[R]}_{n})$ to automatically target $p_{n}(\x_{[n]})$. 
%$\gamma^{(i)}$ can be in this case increased as the particles in the high probability density region for $p_{n+1}(\x_{[n]})$ are copied and can further explore it thanks to the transition kernel that is applied to them.
In Section \ref{sec:rbm}
we experimented with two alternative variations of ARM.
The variations are different mechanisms to generate a new set targeting $p_{n}(\x_{[n]})$:

\paragraph{ARM-anticipate}
As done in ARM, this method starts by copying the old set of particles, $(\x^{[R]}_{[n]}, \widetilde{w}^{[R]}_{n})$.
Before moving this new set, more copies of the particles are made, so that $p_{n+1}(\x_{[n]})$ is better approximated.
This anticipates the information given by $p_{n+1}$. 
Borrowing an idea from residual resampling \citep{LiuChenSMC},
these more promising particles can be, for instance, those whose indexes are in
$\Lcal \defined \{ r : R \lfloor \widetilde{w}^{r}_{n+1} \rfloor > 0 \}$.
Particles with indexes in $\Lcal$ can be split into 
$N_r = \lfloor R \widetilde{w}^{r}_{n+1} \rfloor$ copies,
and their weights set to $\widetilde{w}_{r}^{n} / N_r$ to ensure that the new set still targets
$p_{n}(\x_{[n]})$. Particles with indexes not in $\Lcal$ are kept as they are. A transition kernel that leaves $p_n(\x_{[n]})$ invariant is then applied to them.
The total number of created particles will be $S=R+\sum_r N_r$.

\paragraph{ARM-reseed}
New particles $\x^{[S]}_{[n]} \sim p_{n}(\x_{[n]})$
may be generated from any other sampler, like an MCMC algorithm, weighted with $\widetilde{w}_{n}^{[S]} = 1 / S$,
and added to the old set of particles.
The motivation is that the newly added particles are completely independent from the current set,
and may therefore be from high density regions in $p_n$ that were approximated poorly before,
possibly allowing a significant increase in $\gamma^{(i)}$. 
This method is entirely application-specific, and running
the new sampler to convergence to obtain $\x^{[S]}_{[n]}$ might be a costly operation.

%!TEX root = main.tex

\section{Gaussian Process Classification} \label{sec:gpc}

As a first evaluation of ARM,
we consider a Gaussian Process (GP) classification model (GPC), where data annealing is used to construct the sequence of distributions.
A GP specifies a prior distribution on functions $x : \bxi \to \mathbb{R}$, so that its values $x_n = x(\bxi_n)$ are correlated through a prior covariance matrix $\K$ that depends on the inputs $\bxi_n$.
In a GPC model an observed class label $y_n \in \{ -1, +1 \}$ depends on $x_n$
through a likelihood $p(y_n | x_n) = \Phi(y_n x_n)$,
the probit link function being $\Phi(x) = \int \Theta(z) \Ncal(z ; x, 1) \, \drm z$.
The step function $\Theta(z)$ is one if its argument is nonnegative, and zero otherwise.
Using the step function, the joint model is
\begin{align*}
p(\y, \z, \x)
& = p(\y | \z) \, p(\z | \x) \, p(\x) \\
& = \prod_{n = 1}^{N} \Theta(y_n z_n) \, \Ncal(z_n ; x_n, 1) \cdot \Ncal(\x ; \0, \K) \ ,
\end{align*}
and we are interested in the marginal likelihood $Z = p(\y)$
as a function of $\K$.
Two representations of $Z$ arise from either integrating out $\z$ to give
\begin{equation} \label{eq:gpc-x}
p(\x | \y) = \frac{1}{p(\y)} \exp \left( -\frac{1}{2} \x^T \K^{-1} \x
+ \sum_{n} \log \Phi(y_n x_n) + c_1 \right) 
\end{equation}
with $c_1 = \frac{N}{2} \log 2 \pi - \frac{1}{2} \log |\K|$,
or integrating out $\x$ to yield
\begin{equation} \label{eq:gpc}
p(\z | \y) = \frac{1}{p(\y)} \exp \left( -\frac{1}{2} \z^T (\K + \I) ^{-1} \z
+ \sum_{n} \log \Theta(y_n z_n) + c_2 \right) \ ,
\end{equation}
where $c_2 = \frac{N}{2} \log 2 \pi - \frac{1}{2} \log |\K + \I|$.
Uncertainty is shifted from the likelihood to the prior between these two representations.

\subsection{Implementation}

ARM can be implemented using either the formulation in \eqref{eq:gpc-x} or in \eqref{eq:gpc}.
There are subtle differences between their MCMC transition kernels in the move-step.
The kernel could be a Gibbs sampler for each variable in $p(x_i | y_i, \x_{[n] \wo i})$
or in $p(z_i | y_i, \z_{[n] \wo i})$, where $\wo$ is read as ``without''.
The Gibbs sweeps for $i=1,\ldots,n$ are parameter-free, and for the number of particles under consideration,
computationally much faster
than kernels that make use of gradient information. 

The representation in \eqref{eq:gpc} allows for a more efficient sampler in the move-step than \eqref{eq:gpc-x}
(see \ref{sec:gpc-appendix}).
The parameter-free Gibbs sweep has variances either $1 / [\K^{-1}]_{ii}$ for (\ref{eq:gpc-x}), or $1 / [(\K+\I)^{-1}]_{ii}$ in the case of (\ref{eq:gpc}).
If we introduce the scaling of the covariance function in $\K= \sigma^2 \K_0$, it is easy to see that the variance of the Gibbs sampler scales with $\sigma^2$ for both formulations when $\sigma^2 \gg 1$, but for $\sigma^2 \ll 1$ the variance scales with $\sigma^2$ for (\ref{eq:gpc-x}) and is constant
for (\ref{eq:gpc}).
Gibbs sampling from (\ref{eq:gpc}) is thus more widely applicable as the step-size will in general be larger. Furthermore, the 
representation (\ref{eq:gpc}) has the additional advantage as it is amenable to fast \emph{slice sampling}
and avoids the computation of inverse Gaussian cumulative density functions $\Phi^{-1}$.
The details of all the steps necessary to implement ARM for GPC are given in \ref{sec:gpc-appendix}.

\subsection{Experimental results}

\begin{figure*}[t!]
%\vspace{-0.1cm}
\begin{center}
\includegraphics[width=0.49\textwidth]{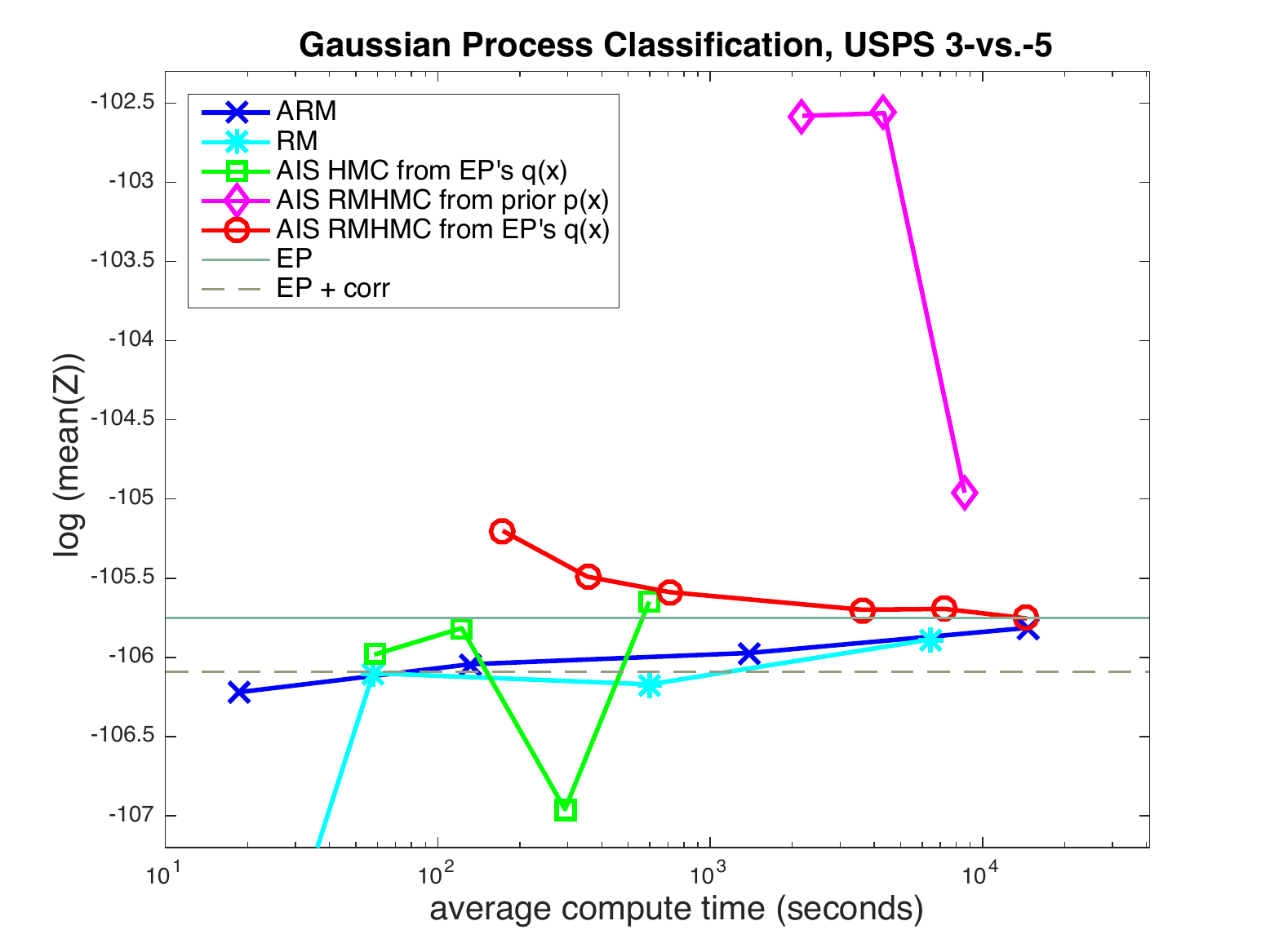}
\includegraphics[width=0.49\textwidth]{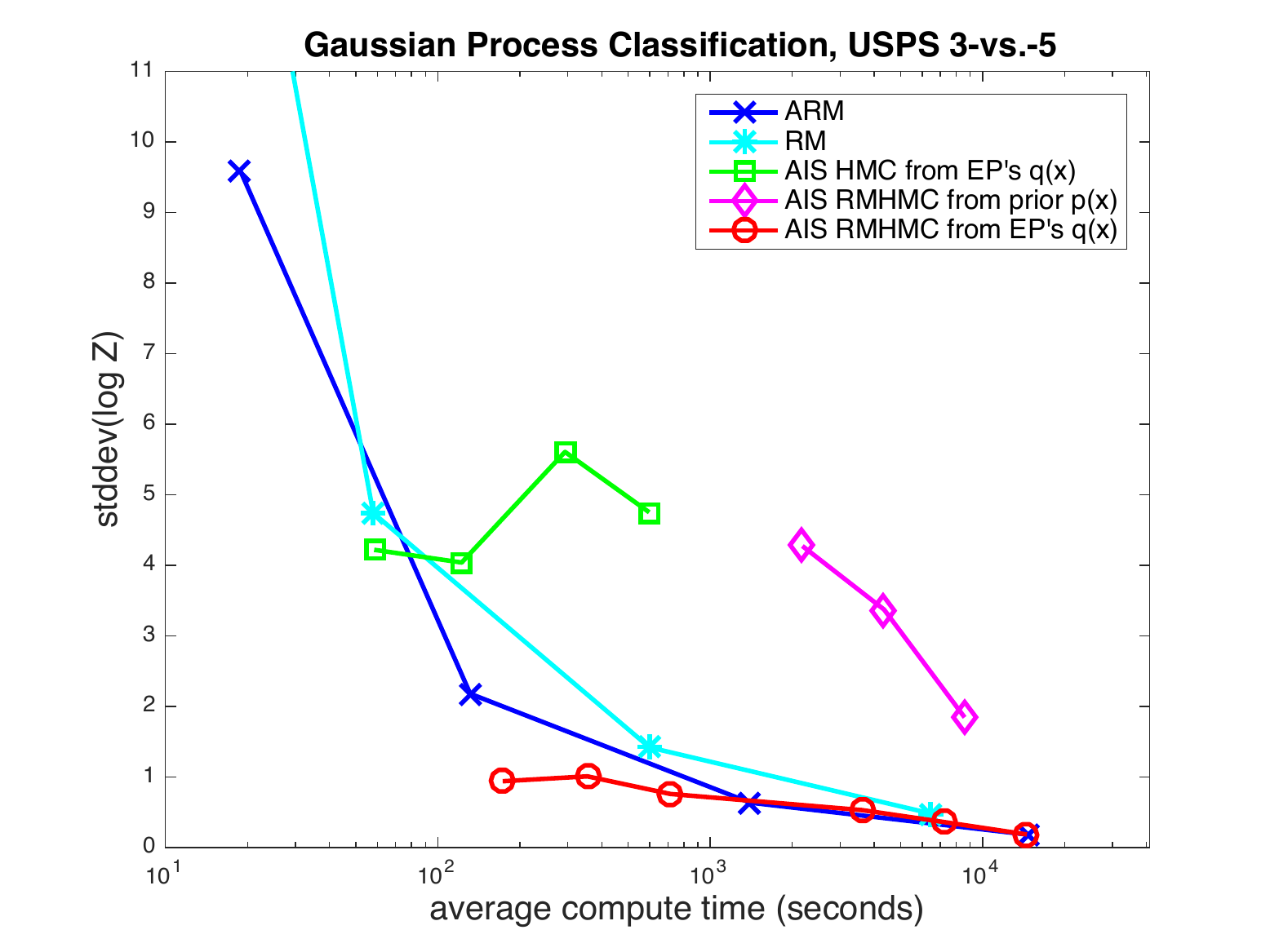}
\caption{
An extensive time-based comparison for GPC on the USPS 3-vs.-5 data set, using a highly correlated prior.
The \emph{left} plot shows the log mean of 80 $\widehat{Z}$ Monte Carlo estimates for each method setting
(the mean of 80 $\widehat{Z}$'s, in the log domain),
while the right plot shows the standard deviation of the 80 $\log \widehat{Z}$ estimates.
Without having to piggyback on a good deterministic EP approximation $q(\x)$, the closest alternative to ARM and RM
is AIS that anneals between the prior $p(\x)$ and the posterior using a RHMHC sampler
(implementation details in \cite{paquet16riemannian}).
The \emph{left} plot shows various discretisation or ``finite particle'' biases: AIS discretises a continuous $\beta \in [0,1]$
interval, and here approaches $\log Z$ from above (see \ref{ap:tempering} for a discussion on the discretization bias).
In this example, the average ARM and RM estimates approach $\log Z$ from below.
The dotted line indicates EP's approximation, and the solid line a second order correction to the EP solution \cite{OpperPerturbative}.
}
\label{fig:gpc}
\end{center}
% \vskip -0.2in
\end{figure*}

We evaluate the efficiency of ARM on the USPS 3-vs.-5 data set \cite{KussAssessing},
using a covariance function $K_{mn} = k(\bxi_m, \bxi_n) = \sigma^2 \exp(-\frac{1}{2} \| \bxi_m - \bxi_n\|^2 / \ell^2 )$ that correlates inputs $\bxi_m$ and $\bxi_n$ through a length scale $\ell = \exp(4.85)$ and amplitude parameter $\sigma = \exp(5.1)$.\footnote{ARM was evaluated on \cite{KussAssessing}'s entire $(\log \ell, \log \sigma)$-grid, of which this setting proved to be the hardest.}
Our main basis for comparison is Annealed Importance Sampling (AIS) \cite{NealAIS} using
different versions of Hamiltonian Monte Carlo (HMC) methods for the transition kernel.\footnote{The posterior density
is very correlated. The mixing rate of a single Gibbs sampler
was too slow in our simulations to get a competitive estimate of the normalizing constant, when used in conjunction with AIS.}
Such a highly correlated high-dimensional prior highlights some deficiencies in
a basic HMC method, where mixing can be slow due to a sample's leapfrog trajectory oscillating
up and down the sides of a valley of $\log p(\y | \x)^{\beta} p(\x)$, without actually progressing
through it.
In order to get a working HMC sampler for the problem, we derived a Riemannian Manifold HMC
method (RMHMC) \cite{RMHMC} for GP models.
To our knowledge, this has not been done before, and as it would be of independent interest,
detailed pseudo-code is given in \cite{paquet16riemannian}.
To further aid AIS with different HMC methods, we additionally let AIS anneal from
a Gaussian approximation $q(\x)$ to the GPC posterior, instead of the prior.
The approximation $q(\x)$ was obtained with Expectation Propagation (EP).

To test the importance of the the MCMC kernel, we also compared the proposed method against a more basic SMC algorithm with no move-step, using a high number of particles that matched the computational budget of ARM. However, the estimates obtained with this sampler were
much worse than the ones obtained with ARM, and are therefore not included in the following analysis.

Figure \ref{fig:gpc} compares the estimates of $\log Z$ obtained with ARM and AIS to the required computation time. Broadly, we see that ARM makes better use of a computational budget than RM.
Secondly, as the \emph{only} competitive versions of AIS with HMC have to rely on outside information through $q(\x)$, SMC methods, in the spirit of ``hot coupling''  \citep{Hamze05hotcoupling},
are unequivocally better workhorses for estimating normalizing constants in this context.
The details of the methods are:

\begin{description}
\item[ARM] in uses $i_{\max} = 50$
with $R_n^{(0)}=R = 10^2, 10^3, 10^4, 10^5$ respectively. Residual resampling is done when $R_{\mathrm{eff}} < 0.9 R$.
\item[RM] uses one Gibbs sweeps in each move-step, with $R = 10^2, 10^3, 10^4, 10^5$,
and does residual resampling only when $R_{\mathrm{eff}} < 0.9 R$.
The mean and standard deviations for runs with $R = 10^2$ are -110.2 and 20.9 and are trimmed from the plots.
% Marco, this is why we originally did two sweeps!
% The choice of two sweeps is made to approximately match the variance of ARM in the estimate of the normalizing constant, and it is empirically equivalent to using more particles with only 1 sweep.
%%
\item[AIS HMC from $q(\x)$] runs AIS from the EP's $q(\x)$ at $\beta=0$ to $p(\x | \y)$
at $\beta = 1$ using intermediate distributions
\begin{equation} \label{eq:aisProb}
p_{\beta}(\x) = \frac{1}{Z(\beta)} \left( \prod_n \Phi(y_n x_n) \cdot {\cal N}(\x ; \0, \K) \right)^{\, \beta} q(\x)^{1 - \beta} \ .
\end{equation}
Note that a starred label indicates that the estimates were aided by $q(\x)$.
A HMC transition kernel with $l_{\max} = 200$ \emph{leapfrog} steps
is used at each $\beta \in [0,1]$ value.
AIS's $\beta$-grid is a geometric progression (geometric discretization; see \cite{kofke2002acceptance}) over $B = 10^3,
2 \times 10^3, 5 \times 10^3$ and $10^4$ $\beta$-values, and these constitute the four green squares in
Figure \ref{fig:gpc}.
A step size $\epsilon = 0.02$ was used per proposal, and both $l_{\max}$ and $\epsilon$ were carefully tuned to the problem.
The simplest AIS-HMC version, which anneals from $p(\x)$ and not $q(\x)$, didn't obtain estimates inside the bounds of Figure \ref{fig:gpc}, and is excluded. 
\item[AIS RMHMC from $p(\x)$] anneals from $p(\x)$ to $p(\x | \y)$ using
%%%
\begin{equation} \label{eq:aisProbP}
p_{\beta}(\x) = \frac{1}{Z(\beta)} \left( \prod_n \Phi(y_n x_n) \right)^{\beta} {\cal N}(\x ; \0, \K) \ ,
\end{equation}
and replaces HMC with a more advanced RMHMC
that uses $\epsilon = 0.1$ and $l_{\max} = 10$ leapfrog steps per proposal at each $\beta$ value.
The inverse temperature $\beta$ was geometrically discretized to $B = 500, 10^3$ and $2 \times 10^3$ values.
Notice that due to the overhead of simulating Hamiltonian dynamics on a Riemannian manifold,
the $\beta$-interval is less discretized than for \emph{AIS HMC}.
This method is further described in the online supplementary material \cite{paquet16riemannian}.
\item[AIS RMHMC from $q(\x)$] anneals from $q(\x)$ to the posterior (see Equation \ref{eq:aisProb}) 
using a RMHMC kernel
($\epsilon = 0.1, l_{\max} = 10$). The $\beta$-interval is geometrically
discretized using $B = 25, 50, 100, 500, 10^3$ and $2 \times 10^3$ points.
\end{description}

It is known that the EP estimate of $\log Z$ is remarkably accurate for this problem \cite{KussAssessing}, hence EP's $\log Z$ estimate and its a second-order corrected estimate
\cite{OpperPerturbative} are given for reference.

Our last observation is a practical one. ARM and RM are simple and tend to be more robust
than AIS with HMC or Metropolis-Hastings, as they have little dependence on external parameters.
HMC, on the other hand, relies on carefully tuned settings of $\epsilon$ and $l_{\max}$,
or requires more complicated extensions like RMHMC used here, or approaches like the No-U-Turn sampler \cite{HoffmanNoUTurn}.

%\todo{
%Although approaches like the No-U-Turn sampler \cite{HoffmanNoUTurn}, in conjunction with Nesterov's dual averaging scheme
%\cite{Nesterov}, . It is empirically underscored by Figure \ref{fig:gps-vs-ais},
%and the discussion surrounding Figure \ref{fig:rbm_comp} for RBMs.
%}
%
%\todo{Results for R=1e2,1e3 and 1e4 are attached. Overall they are significantly better than what we had before, I think.}
%\todo{
%I have modified the number of sweeps used in our approach to 
%nsteps = sqrt( par.corr * nparticles / eff-samp ) + 1;
%where par.corr=50. This is because the theoretical formula assumes independent samples. The number of independent samples should roughtly be divided by the correlation length which I for now set to 50. Next up is an actual calculation of it based upon measuring the correlation length of the weights (or log weights) empirically. }

%!TEX root = main.tex

\section{Restricted Boltzmann Machines} \label{sec:rbm}

A Restricted Boltzmann Machine
(RBM) is a bipartite binary graphical model, connecting visible units  $\x \in \{ 0, 1\}^N$ to hidden binary units $\h \in \{ 0, 1 \}^H$ through
\[
p(\x,\h) = \frac{f(\x,\h)}{Z} = \frac{1}{Z} \exp({\x^T\W\h+\a^T\x+\b^T\h}) \ .
\]
The $N \times H$ weight matrix $\W$ defines the connections between the two layers, $\a$ is a $N \times 1$ bias term relative to the visible units, and $\b$ is a $H \times 1$ bias term for the hidden units. 

To sequentially form an RBM, we can start with a graph containing only the hidden units, and keep adding a new visible unit (with corresponding weights and bias) at each iteration; see Figure \ref{fig:rbm}.
Instead of working with the joint distribution of visible and hidden units, it is more convenient to have the latter summed out:
\begin{align}\label{eq:pxnRBM}
p_n(\x_{[n]}) = \frac{1}{Z_n} e^{{\a_{[n]}^T \x_{[n]}}}\prod_{h=1}^H \left( 1+ e^{{b_h+\W_{[n],h}^T\x_{[n]}}} \right)  \ .
\end{align}

\tikzstyle{state}=[shape=circle,draw=black]
\tikzstyle{txt}=[shape=circle]
\tikzstyle{lightedge}=[dashed]
\tikzstyle{mainedge}=[thick]
\tikzstyle{visstate}=[state,thick,fill=black!20,,inner sep=0pt,minimum size=0.7cm]
\tikzstyle{hidstate}=[state,dashed,scale=0.9]
\begin{figure}[!t]
\begin{center}
\begin{tikzpicture}[scale=0.55]
% 1st model
\node[visstate] (s1) at (0.7,2) {$h_1$};
\node[visstate] (s2) at (0,0) {$x_1$}
edge[mainedge] (s1);
\node[visstate] (s3) at (2.1,2) {$h_2$}
edge[mainedge] (s2);
\node[hidstate] (s4) at (1.4,0) {$x_2$}
edge[lightedge] (s1)
edge[lightedge] (s3);
\node[hidstate] (s5) at (2.8,0) {$x_3$}
edge[lightedge] (s1)
edge[lightedge] (s3)
;

% arrow
\node[txt] (a1) at (4,1) {$\Rightarrow$};

% 2nd model
\node[visstate] (s1) at (5.2+0.7,2) {$h_1$};
\node[visstate] (s2) at (5.2+0,0) {$x_1$}
edge[mainedge] (s1);
\node[visstate] (s3) at (5.2+2.1,2) {$h_2$}
edge[mainedge] (s2);
\node[visstate] (s4) at (5.2+1.4,0) {$x_2$}
edge[mainedge] (s1)
edge[mainedge] (s3);
\node[hidstate] (s5) at (5.2+2.8,0) {$x_3$}
edge[lightedge] (s1)
edge[lightedge] (s3)
;

% arrow
\node[txt] (a1) at (5.2+4,1) {$\Rightarrow$};

% 3rd model
\node[visstate] (s1) at (5.2+5.2+0.7,2) {$h_1$};
\node[visstate] (s2) at (5.2+5.2+0,0) {$x_1$}
edge[mainedge] (s1);
\node[visstate] (s3) at (5.2+5.2+2.1,2) {$h_2$}
edge[mainedge] (s2);
\node[visstate] (s4) at (5.2+5.2+1.4,0) {$x_2$}
edge[mainedge] (s1)
edge[mainedge] (s3);
\node[visstate] (s5) at (5.2+5.2+2.8,0) {$x_3$}
edge[mainedge] (s1)
edge[mainedge] (s3)
;
\end{tikzpicture}
\end{center}
\vspace*{-0.4cm}
\caption{Sequential formation of a simple RBM.}
\label{fig:rbm}
\end{figure}
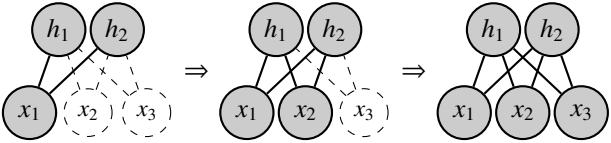

The initialization of the ARM algorithm is
straight-forward, as it is easy to sample from $p_{[1]}(x_{[1]};a_{[1]},\b,\W_{[1],[H]})$ and compute $Z_1$.
%Note that the hidden biases $\b$ are introduced in the model from the very beginning:
%at iteration $n$ we just add $a_{n+1}$ and the weights $\W_{n+1,:}$ that connect the $(n+1)$-th visible unit to all the hidden units.
In the {move}-step, a possible parameter-free transition kernel is the standard Gibbs sampler in which first we sample $\h|\x$ and then $\x|\h$ in parallel \citep{cd}. To improve mixing,
one may repeatedly apply the transition kernel (we used $10$ iterations in our experiments).
The weight updates used in the {smooth}-step are
%$$
%\mathcal{W} \left( \x_{[n]} \right) =\sum_{x_{n+1}} \left( \exp(a_{n+1}x_{n+1})\prod_{h=1}^H \frac{1+\exp\left( b_h+\W_{[n],h}^T \x_{[n]}^r+W_{n+1,h} x_{n+1} \right)}{1+\exp\left( b_h+\W_{[n],h}^T \x_{[n]}^r \right)} \right) \ ,
%$$
\begin{equation} \label{eq:upd_rbm}
\mathcal{W} \left( \x_{[n]}^r \right) =\sum_{x_{n+1}}  e^{a_{n+1}x_{n+1}}\prod_{h=1}^H \frac{1+e^{ g_{n,h}^r+W_{n+1,h} x_{n+1} }}{1+e^{ g_{n,h}^r }}  \ ,
\end{equation}
where $g_{n,h}^r \doteq b_h+\W_{[n],h}^T \x_{[n]}^r$.
The two terms in the sum in \eqref{eq:upd_rbm}, normalized by $\mathcal{W} ( \x_{[n]}^r )$, form $p(x_{n+1} | \x_{[n]}^r)$ in the {augmentation}-step.
%An application-specific method for the {generate}-step will be introduced in the following.

\subsection{Experimental results} \label{sec:rbm_res}

\begin{figure*}[t]
\begin{center}
\includegraphics[width=0.48\textwidth]{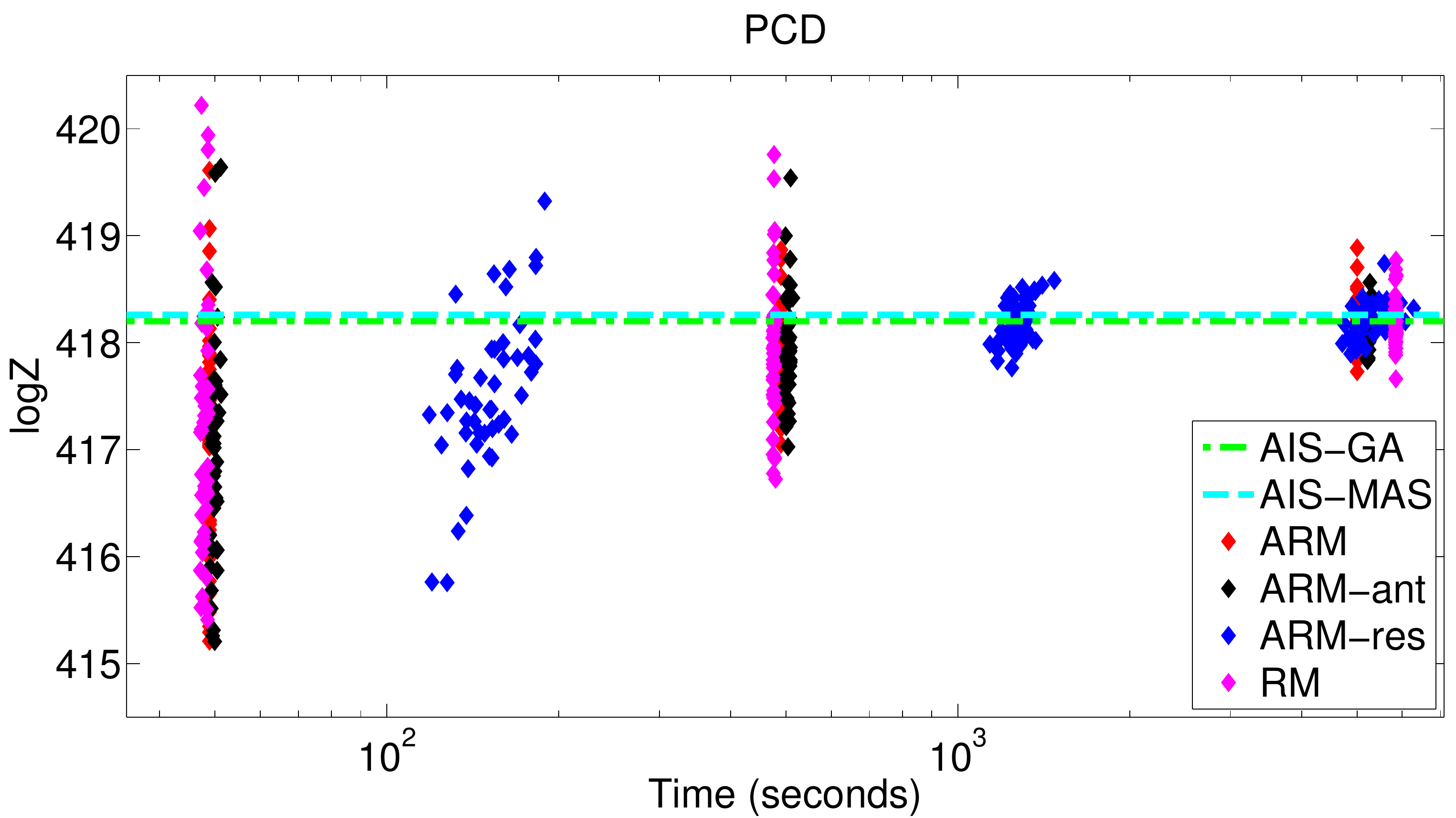}\hfill
\includegraphics[width=0.48\textwidth]{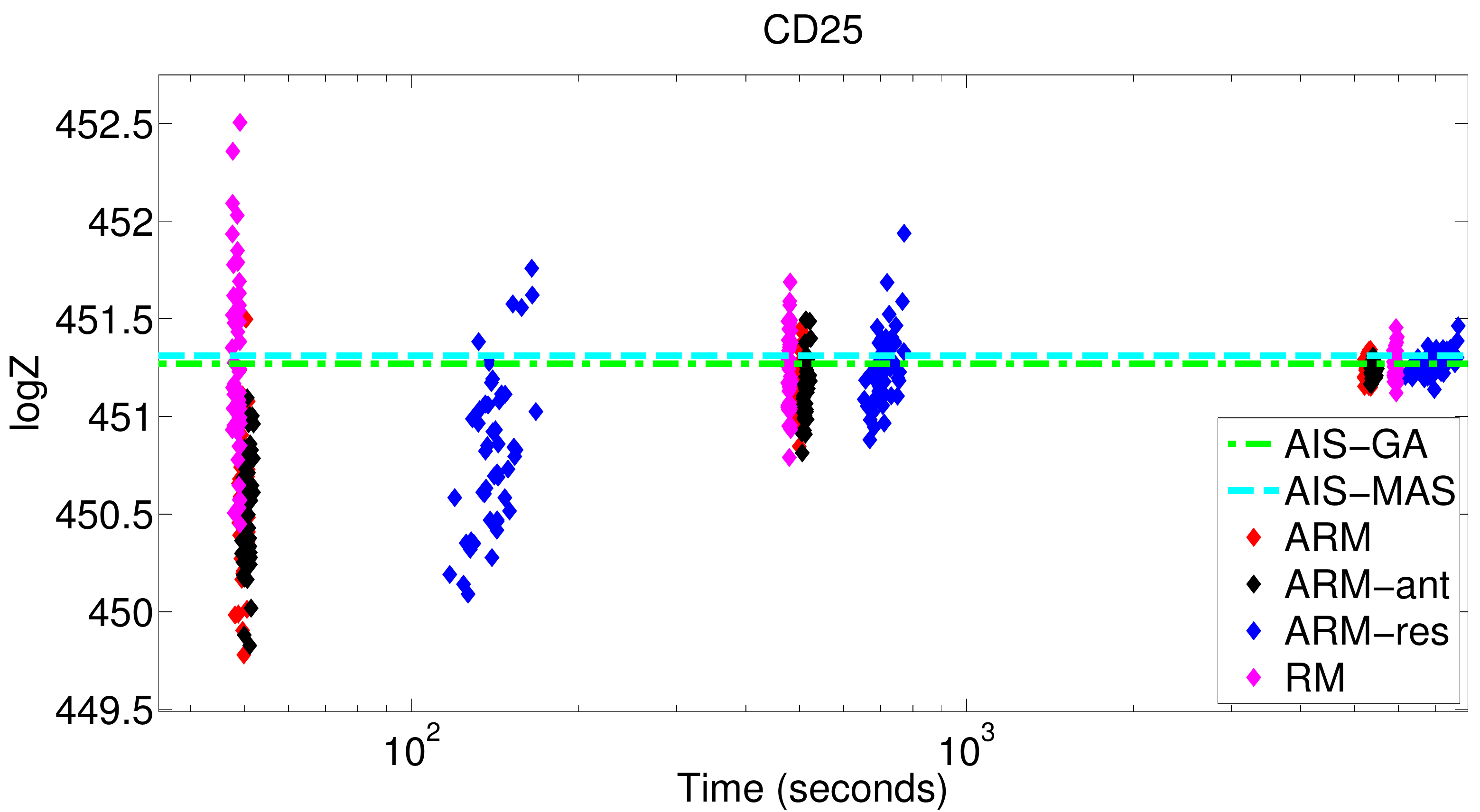}
\caption{Comparison between ARM, RM and AIS for two different RBMs, labelled as PCD and CD25.
\textit{For both models}: in all the simulations we used $t=10$ Gibbs steps for the transition kernel of the move-step, $\gamma_{\rm thr}=0.7$, ARM and ARM-anticipate used $R = 10^2, 10^3, 10^4$ particles and $i_{\max}=3$,
RM used  $R = 10^2, 10^3, 1.2 \times 10^4$ particles.
\textit{For PCD}: ARM-reseed used $R = 10^2, 10^3, 6 \times 10^3$ particles, $i_{\max}=2$ and 3000 Gibbs transitions when generating new particles.
\textit{For CD25}: ARM-reseed used $R = 10^2, 10^3, 10^4$ particles, $i_{\max}=2$ and 1500 Gibbs transitions when generating new particles.
The results for AIS were extracted from \citep{AISmoments}, and were obtained with 5000 chains, $10^5$ intermediate distributions and 1 Gibbs transition for each intermediate distribution.}
\label{fig:rbm_res}
\end{center}
\end{figure*} 

We compare the performance of ARM on the two most difficult RBM models used in \cite{AISmoments}.
Both models were trained on the MNIST handwritten digits dataset \cite{mnist}, the first one with persistent contrastive divergence (PCD) \cite{pcd} and the second one with contrastive divergence \cite{cd} with 25 steps of Gibbs sampling (CD25). The RBMs have 784 visible units and 500 hidden units, making the exact computation of $Z$ intractable.
In \cite{AISmoments} the partition function is estimated with AIS, using a path based on averaging the moments of the initial and target distribution instead of the usual geometric one. 
The algorithm presented is computationally very expensive: first the moments of the target distributions are estimated using $10^3$ independent Gibbs chains with 11000 Gibbs steps each, then the parameters of 9 intermediate RBMs have to be fit in order to match the averaged moments at 9 different temperatures (knots of a spline), and finally a geometric path with $10^4$ intermediate distributions is used in order to pass from one RBM at one knot to the next one,
therefore giving $K = 10^5$ intermediate distributions in total.
The best performing initial distribution used for AIS in \cite{AISmoments} is the base rate RBM, in which the visible biases are set to the average pixel values in the MNIST training set and all the other parameters are set to 0.
In a similar way information on the training data can be exploited by ARM as well: units are introduced starting from the most active ones (i.e.~those whose variance in the training set is higher) so that higher density regions are explored from the very beginning.

\begin{figure}[!t]
\begin{center}
\begin{tikzpicture}[node distance = 0.25\textwidth, auto]
%\node[align=center] (400) 
%    {200 visible units\\ \includegraphics[width=0.2\textwidth]{figures/digit_200_10}};
\node[align=center] (600) 
    {200 visible units\\ \includegraphics[width=0.2\textwidth]{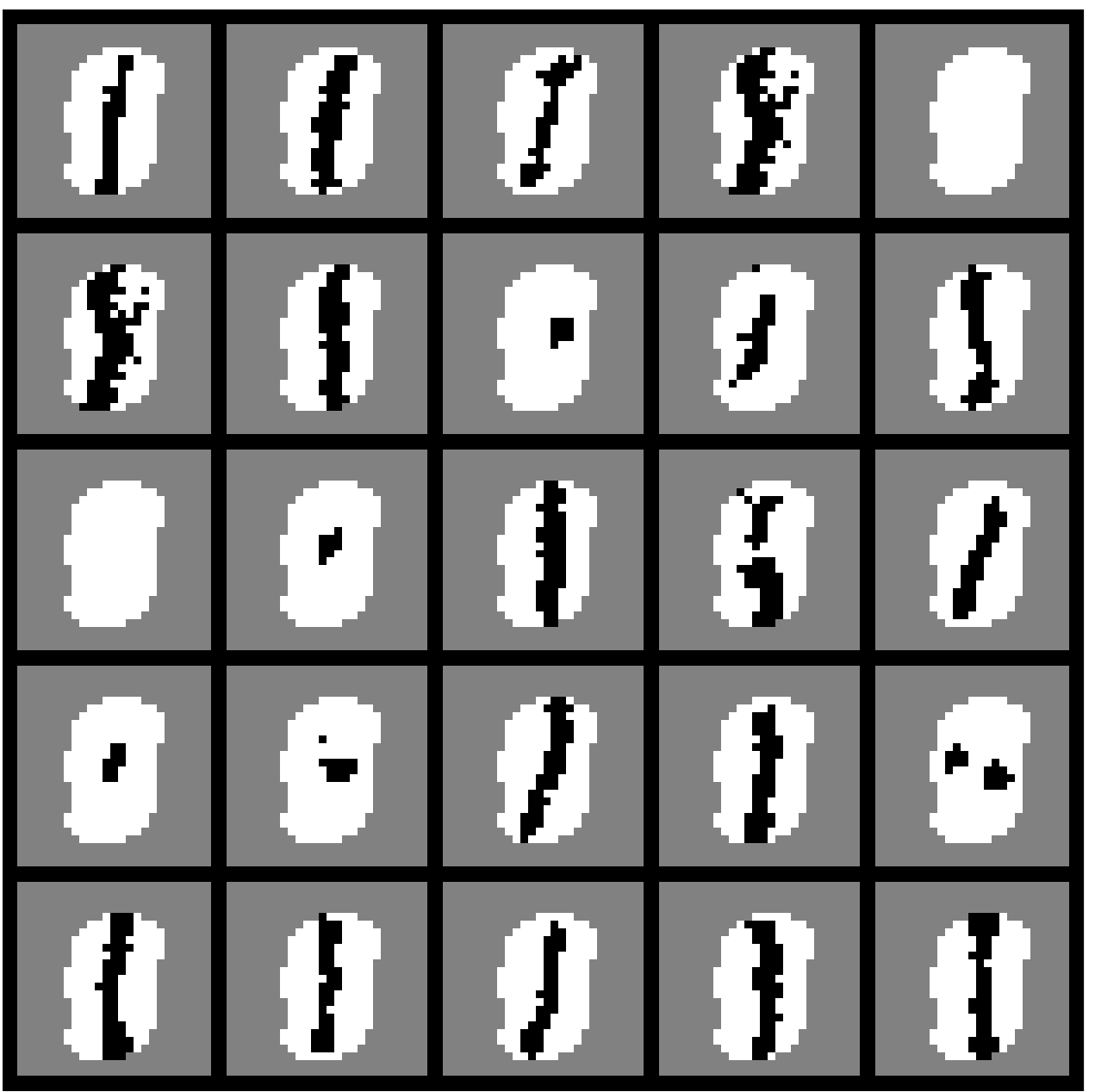}};
%\node[align=center,right of = 400] (600) 
%    {550 visible units\\ \includegraphics[width=0.2\textwidth]{figures/digit_550_1}};
\node[align=center,right of = 600] (784) 
    {784 visible units\\ \includegraphics[width=0.2\textwidth]{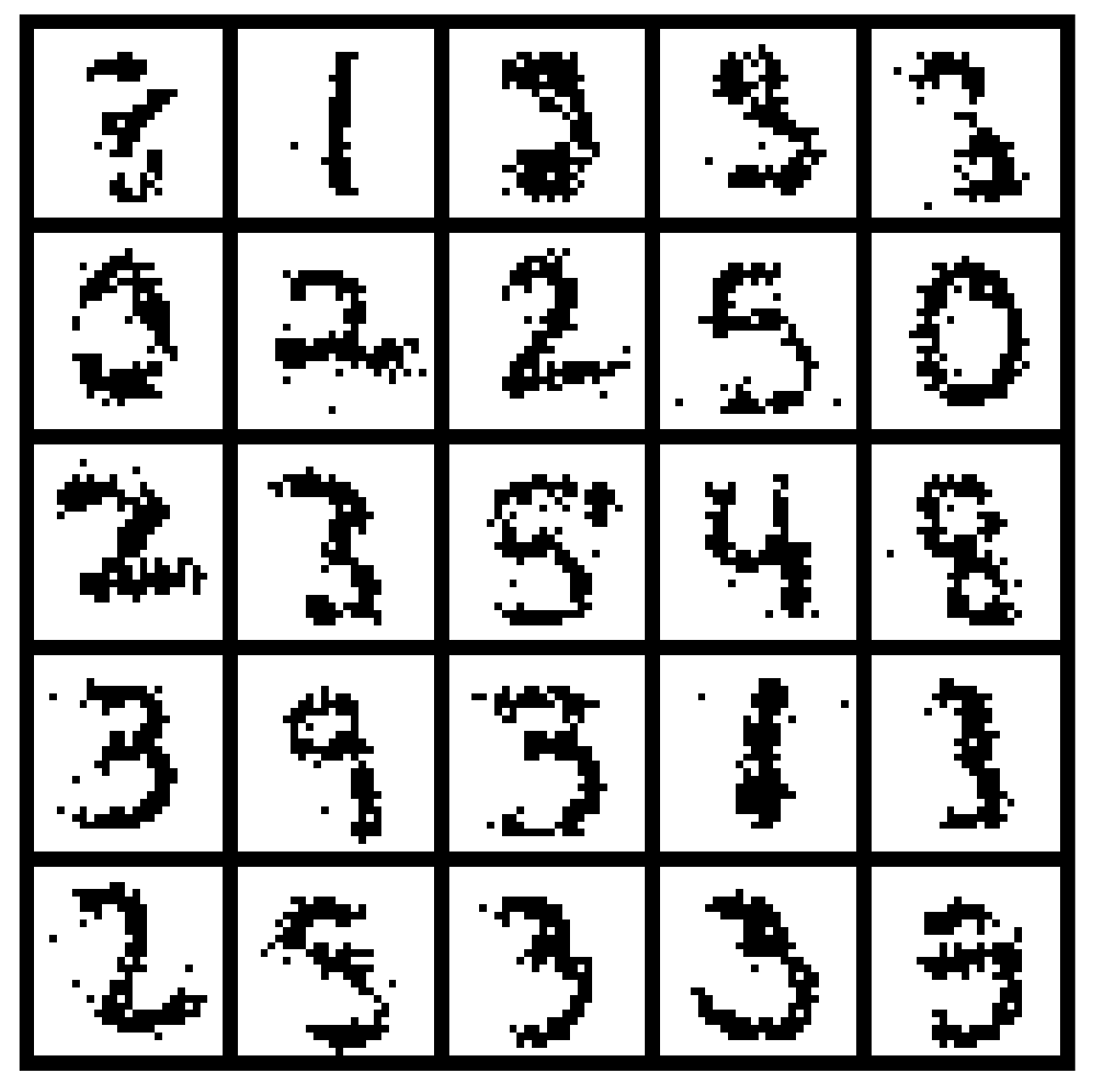}};
%\draw[->,thick,line width=0.8mm,draw=blue] (400) -- (600) {};
\draw[->,thick,line width=0.8mm,draw=blue] (600) -- (784) {};
%\path[draw=blue,solid,line width=1mm,fill=black, preaction={-triangle 90,thin,draw,shorten >=-1mm}] (200) -- (400);
%\path[draw=black,solid,line width=1mm,fill=black, preaction={-triangle 90,thin,draw,shorten >=-1mm}] (400) -- (600);
%\path[draw=black,solid,line width=1mm,fill=black, preaction={-triangle 90,thin,draw,shorten >=-1mm}] (600) -- (784);
\end{tikzpicture}
\end{center}
\vspace*{-0.5cm}
\caption{Example of particles approximating $p_n(\x_{[n]})$ using the RBM model trained with PCD. Black and white pixels represent ones and zeros, while excluded visible units are colored gray.}
\label{fig:digits}
\end{figure}

Figure \ref{fig:digits} shows some of the particles that were generated by ARM at iteration $n=200$ and $n=784$.
We see that the visible units (pixels of the image in this case) are sequentially added,
and when $n$ is high enough the particles start looking like real handwritten digits digits.

The particle set can be extended with $S$ new independent particles $\x^{[S]}_{[n]}$ using  ARM-reseed (see Section \ref{sec:gener}).
To ensure that the new set of particles targets $p_{n}(\x_{[n]})$, a few thousand steps of the same Gibbs kernel of the move-step are applied to $S$ examples sampled randomly from the MNIST training set.
To minimize the computational overhead of this costly step, the actual number of new particles generated at iteration $i$ was $S=\max\left[(1-\gamma^{(i)}/\gamma_{\rm thr})R,100\right]$, where $R$ is the baseline number of particles. This means that if the current set of particles is already a good approximation of the distribution of interest, only few new particles are created.
At least 100 particles were generated at each step, so that the new set provides a sufficiently good approximation to $p_{n}(\x_{[n]})$. 

The results from 50 runs of ARM and its two variants from Section \ref{sec:gener} are shown
in Figure \ref{fig:rbm_res}, using an increasing number of particles.
As a reference we also show the results for AIS from \citep{AISmoments},
which were obtained with a geometric averages (AIS-GA) path and a moment averages spline (AIS-MAS) path,
using $K=10^5$ intermediate distributions.

Remarkably, ARM allows us to get very close to the results obtained with AIS in less than a minute of computation time.
Most importantly, for ARM the tuning of the parameters was almost effortless, as these are merely a function of the allowed time budget.
By considering the estimates in Figure \ref{fig:rbm_res} that were obtained with the highest number of particles
(the rightmost estimates in each plot),
we notice that ARM reduces the variance of RM estimates using less computational power.
ARM-anticipate, which creates new particles in a ``smarter'' way, outperforms simple ARM in terms of variance of the estimates.
For the RBM trained with PCD, generating a new set of independent particles with ARM-reseed significantly improves
the efficiency of the sampler; see in particular the results obtained with a baseline of $R=10^3$.
On the other hand,
other methods give comparable results to ARM-reseed in less time for the CD25-trained RBM.
As noted in \cite{AISmoments}, PCD seems to be a more difficult model to sample from, and as such a completely new set of particles could be beneficial.
%Most importantly, for ARM the tuning of the parameters was effortless: the number of particles, the number of Gibbs steps per iterations, the maximum number of iteration, and the threshold for ESS have to be specified, and \emph{a priori} these are merely a function of the allowed time budget.

We can get an insight into the computational performance of both ARM and AIS by comparing the efficiency of the respective Gibbs kernels, as they are by far the most time consuming operation in both algorithms.
One particle of ARM essentially corresponds to one run of AIS.
At iteration $n$ of ARM, the Gibbs sampler has complexity $\mathcal{O}(tHn)$, where $t$ is the number of Gibbs steps used. As $n=1,\ldots,N$, the overall complexity is $\mathcal{O}(tHN^2)$.
The complexity of AIS (using geometric or moment averages), given $K$ intermediate distributions, is $\mathcal{O}(tHNK)$.
We than see that AIS is much more computationally expensive than ARM, as it typically requires $K \gg N$ to get accurate estimates of normalizing constants (our best performing setups in Figure \ref{fig:rbm_res} use $K = 10^5$ and $N=784$).

%!TEX root = main.tex

\section{Related Work}\label{relWork}

To estimate normalizing constants in smaller scale models, stochastic approximation techniques are the first choice, as they can lead to very accurate results given enough computational time.
%However, if the size of the model allows to use them in a reasonable amount of time, as for the examples shown in this work, they usually represent the primary choice, and will be therefore analyzed more in detail. 
AIS \cite{NealAIS}, used as a comparison in our simulations, is one of the most widely used method for estimating normalizing constants. It belongs to a more general family of methods, known as \textit{tempering methods}, that are based on a one-parameter $\beta\in[0,1]$ extension of the model: $p_\beta(\x) = f(\x,\beta)/Z(\beta)$ such that we interpolate between a usually tractable $p_0(\x)$ and the model of interest $p(\x) = p_1(\x)$.
This was done in \eqref{eq:aisProb}.
The normalizer can then be written as an integral,
\begin{equation} \label{eq:logZbeta}
\log Z - \log Z(0) = \int_0^1 \Ebb_{p_\beta} \left[ \frac{\drm }{ \drm \beta }\log f(\x,\beta)\right] \, \drm \beta \ ,
\end{equation}
that in practice will have to be discretized.
AIS provides in general accurate estimates, but relies on often difficult hand tuning and a high number of intermediate distributions to limit the large variances of the estimate introduced by the discretization of the continuous temperature scale.
A theoretical derivation of this statement can be found in the \ref{ap:tempering}. For standard annealing schemes, such as AIS,
the analysis shows that the spacing between $\beta$s should be roughly $1/\sqrt{V(\beta)}$ with
\[
V(\beta) \defined \Vbb_{p_\beta}\left[ \frac{\drm \log f(\x,\beta)}{ \drm \beta }\right] \ .
\]
The averaging moments annealing algorithm \cite{AISmoments} may be viewed as a scheme for setting the interpolating distributions for exponential families in a way less prone to discretization errors, using moment averages to define the sequence of consecutive distributions rather than the standard choice of geometric averages \citep{salakhutdinov2008}.

Sequential Monte Carlo (SMC) algorithms can take an alternative route to constructing an interpolation scheme to estimate $Z$.
Our choice of sequence is motivated by
a computationally efficient implementation of versions of Hamze and de Freitas's
 \citep{Hamze05hotcoupling}
``hot coupling'' samplers
(see Algorithms \ref{alg:rm} and \ref{alg:rmg}), as well 
a discrete decomposition for $\log Z$ that allows one to estimate it accurately.
With methods such as the ones introduced in \citep{Hamze05hotcoupling} or ARM, a discrete decomposition for the normalizing constant naturally arises; see \eqref{eq:decomposition}.
In the results in Section \ref{sec:gpc} and \ref{sec:rbm} these methods showed superior performance to AIS,
but a more general statement is not possible.
AIS may be a better choice for other models, despite being more difficult to tune
than ARM.
Similarly to our work, \citep{Naesseth2014} shows how the normalizing constant of general probabilistic graphical models can be estimated with sequential Monte Carlo methods by adding the random variables in the graph one by one. 

Other authors have noticed that the performance of sequential Monte Carlo methods can be improved by adapting during inference the sample size. 
\citep{Fox2003} introduces KLD-sampling, that determines the required number of particles so that the Kullback-Leibler divergence between the sample-based maximum likelihood estimate of the posterior approximation in a discretized state space and the sample-based representation of the predictive distribution (used instead of the intractable true posterior) is low enough. 
Similarly, KLD-resampling is introduced in \citep{Li2013}. 
Due to the required discretization of the state space these methods can however only be used in low dimensional applications, such as robot localization and tracking. 
A survey on adaptive resampling techniques for particle filtering can be found in \citep{Li2015}.
 Likelihood-based adaptation, used for example in \citep{Koller1998}, generates new samples until the sum of the unnormalized likelihoods exceeds a predefined threshold. 
In the context of Sequential MCMC for target tracking in large volumes of data, adaptive subsampling of the measurements at each time step can reduce the computational requirements with minor losses in the accuracy of the estimators \citep{Freitas2015,Bardenet2014}.
The Particle Learning approach \citep{Carvalho2010} uses a fully-adapted filter to learn the parameters of a general class of state space models by defining a particle approximation to the joint posterior distribution of states and conditional sufficient statistics for the fixed parameters. Estimation of abruptly time varying parameters can be done with the adaptive approach presented in \cite{Nemeth2014}.
% \todo{The reviewer also suggested \cite{Nemeth2016} as a reference, but it seems loosely related to what we do, I wouldn't put it.. Maybe the reviewer is one of the authors :-)}

\section{Conclusion}\label{conclusion}

In this paper we introduced \textit{Adaptive Resample-Move}, an SMC algorithm that reduces the variance of estimates of normalizing constants by expanding the particle set whenever a better approximation of an intermediate distribution is needed. A theoretical justification for ARM is also given under ideal conditions. Experimental results on two challenging models previously analyzed by other authors (GPC and RBMs), show that despite its simplicity and the minimal tuning required, ARM allows to efficiently find accurate estimates of normalizing constants, and should therefore be considered as a valid alternative to AIS.

\section*{Acknowledgements}
Marco Fraccaro is supported by Microsoft Research through its PhD Scholarship Programme.

%Future research directions can explore other methods to generate the new particles, 

% \section*{References}

\bibliographystyle{elsarticle-num}
\bibliography{sequentialsmoothing}

\appendix

%!TEX root = main.tex

\section{Resample-Move for Gaussian Process classification} \label{sec:gpc-appendix}

In this appendix, we present a MCMC transition kernel for Algorithm \ref{alg:rm}'s move step,
as done in iteration $n$.
The transition kernel performs Gibbs sampling, where each step is performs
numerically fast slice sampling. 
Efficient smooth and augmentation steps are also given.
We repeat (\ref{eq:gpc}) here for $p(\z_{[n]}) = f_n(\z_{[n]}) / Z_n$:
\begin{align*}
p(\z_{[n]} | \y_{[n]}) & = \frac{1}{Z_n} \exp \Bigg( -\frac{1}{2} \z_{[n]}^T (\K_{[n]} + \I_{[n]}) ^{-1} \z_{[n]} \\
& \qquad\qquad\qquad\qquad
+ \sum_{i \in [n]} \log \Theta(y_i z_i) + c \Bigg) \ ,
\end{align*}
% \]
where
$c = \frac{n}{2} \log 2 \pi - \frac{1}{2} \log |\K_{[n]} + \I_{[n]}|$.

\subsection{The move-step}

The move-step in Section \ref{sec:gpc} requires an MCMC transition kernel to
resample $\z_{[n]}^{[R]}$ given its current state.
The move-step draws
samples from $p(z_i | y_i, \z_{[n] \backslash i})$ for
$i \in [n]$ in random order.
Let $\S^{(n)} = \mathbf{\Sigma}_{[n]}^{-1} = (\K_{[n]} + \I_{[n]})^{-1}$ be the inverse prior covariance of
$\Ncal(\z_{[n]} ; \0, \mathbf{\Sigma}_{[n]})$.
The conditional distribution for any $z_i$ is
\begin{equation} \label{eq:gibbsz}
p(z_i | y_i, \z_{[n] \backslash i}) \propto \Theta(y_i z_i) \, \Ncal \left( z_i \, ; \, \mu_i , \, [S_{ii}^{(n)}]^{-1} \right) \ ,
\end{equation}
where the Gaussian has mean
\[
\mu_i = - \sum_{j \in [n] \backslash i} S_{ij}^{(n)} z_j \, \big/ \, S_{ii}^{(n)} \ .
\]
A single Gibbs sweep requires an inner product for $\mu_i$ for each of $n$ samples
with (\ref{eq:gibbsz}),
giving complexity ${\cal O}(n^2)$ for ${\cal K}$.
This inner loop dominates Algorithm \ref{alg:rm}'s cost.

\subsection{Slice sampling in the move-step}

Each Gibbs sample from $p(z_i | y_i, \z_{[n] \backslash i})$ in (\ref{eq:gibbsz})
can be extremely efficiently drawn using a \emph{slice sampler} that only requires two uniform random numbers and the computation of a square root.
We start by drawing height $u \sim u | z_i, y_i$ uniformly between zero and $\exp(-S_{ii}^{(n)}(z_i - \mu_n)^2 / 2)$.
The bounds of the slice are the two roots of the quadratic
$S_{ii}^{(n)}(z_i - \mu_i)^2 + 2 \log u = 0$, possibly clipped at zero according to the sign
of $y_i$. These operations can be concatenated into four steps to update $z_i$
\begin{align}
\textrm{1:} & & v & = \big( (z_i - \mu_i)^2 - 2 [S_{ii}^{(n)}]^{-1} \log (\mathrm{rand}) \big)^{1/2} \nonumber \\
\textrm{2:} & & b_{\mathrm{lo}} & = (\mu_i - v) \, \mathbb{I}[y_i < 0 \ \mathrm{or} \ \mu_i > v ] \nonumber \\
\textrm{3:} & & b_{\mathrm{hi}} & = (\mu_i + v) \, \mathbb{I}[y_i > 0 \ \mathrm{or} \ \mu_i < -v ] 
\nonumber \\
\textrm{4:} & & z_i & = (b_{\mathrm{hi}} - b_{\mathrm{lo}})\mathrm{rand} + b_{\mathrm{lo}} \ , \label{eq:slice}
\end{align}
where $\mathrm{rand}$ produces a ${\cal U}[0,1]$ sample, and $\mathbb{I}[\cdot]$ is one if its argument is true,
and zero otherwise.

\subsection{The smooth- and augmentation steps}

The smooth-step, the conditional density for $x_{n+1}$ for the augmentation-step, as well as
the algorithm's next loop, require $\S^{(n+1)} = \mathbf{\Sigma}_{[n+1]}^{-1}$.
We first expand the inverse with an $\Ocal(n^2)$ operation
\[
\S^{(n+1)} = \begin{pmatrix}
   \S^{(n)} + \s \s^T / \varsigma & \s  \\
   \s^T                           & \varsigma
  \end{pmatrix} ,
\]
using the block matrix invserion
\[
\begin{array}{l}
\varsigma = \left( \Sigma_{n+1, n+1} - \mathbf{\Sigma}_{n+1, [n]}^{T} \S^{(n)} \mathbf{\Sigma}_{[n], n+1} \right)^{-1} \\
\s = - \varsigma \S^{(n)} \mathbf{\Sigma}_{[n], n+1}
\end{array} .
\]
Notation $\mathbf{\Sigma}_{[n], n+1}$ refers to the subvector in $\mathbf{\Sigma}$ that is indexed by rows $[n]$ and column $n+1$.
On obtaining $\S^{(n+1)}$, the smoothing step calculates $w^r$ by averaging the
likelihood for $y_{n+1}$ over a Gaussian conditional distribution $p(z_{n+1} | \y_{[n]}, \z_{[n]})$ with mean and variance
\begin{align*}
m & \defined - \frac{1}{S_{n+1,n+1}^{(n+1)}} \sum_{i=1}^{n} S_{i,n+1}^{(n+1)} z_i^r \\
\sigma^2 & \defined \frac{1}{ S_{n+1, n+1}^{(n+1)} } \ ,
\end{align*}
to yield
\begin{align*}
w^r & = \int \Theta(y_{n+1} z_{n+1}) \, \Ncal( z_{n+1} ;  m , \sigma^2 ) \drm z_{n+1} \\
& = \Phi(y_{n+1} \cdot m / \sigma ) \ .
\end{align*}

\section{Tempering methods}\label{ap:tempering}

Reference \citep{gelman1998simulating} derived an exact asymptotic expression for the bias due to the discretization of
\[
\log Z = \log Z(1) - \log Z(0) = \int_0^1 \Ebb_{p_\beta} \left[ \frac{\drm }{ \drm \beta } \log f(\x,\beta)\right] \, \drm \beta
\]
in \eqref{eq:logZbeta}.
This expression is not computable in practice and the main challenges of tempering methods are to come up with efficient procedures for choosing $f(\x,\beta)$ \cite{AISmoments, gelman1998simulating} and tuning the discretisation of $\beta$ to the specific problem.
Intuitively, a necessary requirement for the successful interpolation is that the intermediate distributions must be sufficiently similar.
In other words, the distribution of the \textit{energy} $\frac{\drm \log f(\x,\beta)}{ \drm \beta }$ for adjacent distributions must be overlapping.
We define $M(\beta)$ as the expectation of the energy,
\[
M(\beta) \defined \Ebb_{p_\beta} \left[ \frac{\drm }{ \drm \beta } \log f(\x,\beta)\right] \ ,
\]
and its change a
$\Delta M(\beta) \defined M(\beta+\Delta \beta) - M(\beta)$.
$\Delta M(\beta)$ should be made of the same order as the fluctuations in energy, $\sqrt{V(\beta)}$ with $V(\beta) \defined \Vbb_{p_\beta}\left[ \frac{\drm \log f(\x,\beta)}{ \drm \beta }\right]$.
Combining therefore $\Delta M(\beta) \approx \sqrt{V(\beta)}$ with $\Delta M(\beta) \approx \frac{\drm M(\beta)}{\drm \beta } \Delta \beta$ we have a yardstick to measure how much we are allowed to change $\beta$:
\[
\Delta \beta \approx \sqrt{V(\beta)} \, \Big/ \, \frac{\drm M(\beta)}{\drm \beta } \ .
\]
We can write  $\frac{\drm M(\beta)}{\drm \beta } = U(\beta)+ V(\beta)$ with $U(\beta) \defined \Ebb_{p_\beta} \left[ \frac{\drm^2 \log f(\x,\beta)}{ \drm \beta^2 }\right]$.
For standard tempering (as used in AIS) we have $\log f(\x,\beta) = \beta \log f(\x) + (1-\beta) \log f_0(\x)$ which gives $\Delta \beta \approx 1/\sqrt{V(\beta)}$.
This result has the simple interpretation that if fluctuations are large we need to use a finer discretization, increasing therefore the computations required. Unfortunately, ``phase transition'' type behaviour, marked by a large increase in fluctuations for a specific $\beta$, may also occur in large statistical models.

\end{document}